\documentclass{article}

\usepackage{iclr2027_conference,times}
\usepackage[utf8]{inputenc}
\usepackage[T1]{fontenc}
\usepackage{amsmath,amsfonts,amssymb,amsthm}
\usepackage{mathtools}
\usepackage{graphicx}
\usepackage{booktabs,multirow}
\usepackage{afterpage}  %
\usepackage{algorithm,algorithmic}
\usepackage{microtype}
\usepackage{xcolor}

\usepackage{amsmath,amsfonts,bm}

\def\eqref#1{equation~\ref{#1}}

\def\1{\bm{1}}

\DeclareMathAlphabet{\mathsfit}{\encodingdefault}{\sfdefault}{m}{sl}
\SetMathAlphabet{\mathsfit}{bold}{\encodingdefault}{\sfdefault}{bx}{n}

\usepackage{hyperref}
\usepackage{url}
\usepackage[capitalize,noabbrev]{cleveref}

\usepackage{xcolor}   %
\usepackage{soul}
\sethlcolor{yellow}

\newcommand{\jiaqi}[1]{}

\usepackage{pifont}
\usepackage[nointegrals]{wasysym}
\definecolor{cmarkcolor}{HTML}{2E7D32}
\definecolor{xmarkcolor}{HTML}{C0504D}
\definecolor{pmarkcolor}{HTML}{C98A1B}
\newcommand{\cmark}{\textcolor{cmarkcolor}{\ding{51}}}
\newcommand{\xmark}{\textcolor{xmarkcolor}{\ding{55}}}
\newcommand{\pmark}{\textcolor{pmarkcolor}{\LEFTcircle}}

\newcommand{\colhead}[1]{\begin{tabular}[b]{@{}c@{}}#1\end{tabular}}
\newcommand{\ourbenchmarkname}{\textsc{FABench}}

\graphicspath{{figs/}}

\title{Benchmarking Forecasting Agents on Yesterday's Tomorrows}

\newif\ifarxiv
\arxivtrue
\ifarxiv
\author{\parbox[t]{\dimexpr\textwidth-2\tabcolsep\relax}{\centering
Yuxuan Wang\textsuperscript{1}, Yishu Wang\textsuperscript{1,2,$\dagger$}, Jiaqi Deng\textsuperscript{1}\\[0.3ex]
{\normalfont \textsuperscript{1}Raven Labs\quad \textsuperscript{2}Zhejiang University}\\[0.3ex]
{\normalfont\small $^{\dagger}$Corresponding author: \texttt{yishu.wang@raven-labs.xyz}}\\[0.9ex]
{\normalfont \raisebox{-0.15em}{\includegraphics[height=1em]{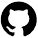}}~GitHub: \href{https://github.com/raven-labs-official/fabench}{github.com/raven-labs-official/fabench}\\
\raisebox{-0.15em}{\includegraphics[height=1em]{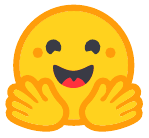}}~Hugging Face: \href{https://huggingface.co/datasets/raven-labs-official/fabench}{huggingface.co/datasets/raven-labs-official/fabench}}}}
\iclrfinalcopy
\else
\author{Anonymous authors\\Paper under double-blind review}
\fi

\begin{document}

\maketitle
\ifarxiv\lhead{Preprint}\fi

\begin{abstract}

Digital agents can now gather and organize information, reason over it, and use tools to complete complex tasks. Forecasting puts all of these abilities to the test: it requires integrating incomplete evidence, inferring causal drivers, and committing to calibrated probabilistic judgments. Existing benchmarks are poorly suited to measuring and improving this ability. Live benchmarks provide feedback only after events resolve, so every evaluation cycle waits on the calendar; fixed-corpus evaluations return results immediately but confine research to pre-collected documents. We introduce \ourbenchmarkname, a scalable backtesting environment for \textbf{F}orecasting \textbf{A}gents. An automated pipeline discovers events, generates questions, and resolves outcomes for any past forecast date, so the benchmark can be renewed to stay ahead of model training cutoffs; its first release poses 287 questions across nine domains. Agents research each question freely in an archive environment that combines an offline archive with live web search screened to prevent temporal leakage. Evaluating agents built on eight recent models, we find that retrieval improves the Brier score of all eight in every retrieval setting and lifts the leading agents to the market consensus on the 54 questions linked to Polymarket, with the best three scoring slightly above it. Research traces reveal recurring failures in how agents weigh and apply evidence, leaving ample room for improvement. We release our code, data, and archive environment as a foundation for community research on forecasting agents.

\end{abstract}

\section{Introduction}
\label{sec:introduction}

Large language models have advanced at a remarkable pace, and the agents built on them now carry out a growing share of everyday knowledge work: writing and debugging code~\citep{jimenez2024swebench}, conducting deep research across many sources~\citep{wei2025browsecomp}, and operating computers through their interfaces~\citep{xie2024osworld}. On some of these tasks, agents now approach the level of human experts~\citep{patwardhan2026gdpval}. A gap remains, however, on complex, long-horizon tasks~\citep{kwa2025measuring} that require an agent to gather large amounts of information, reason about how a situation will unfold, and commit to a decision. Forecasting is the sharpest instance of this challenge. To predict a future event, an agent must integrate incomplete and often conflicting evidence, infer the causal drivers behind it, and compress that understanding into a calibrated probability. It is also a demanding test of decision-making under uncertainty~\citep{tetlock2015superforecasting,mellers2015identifying}: whether an agent can forecast future events as well as human experts, which LLM forecasters have begun to approach~\citep{halawi2024approaching,schoenegger2024silicon,gu2026prophets}.

Progress toward this goal is hard to measure, because existing forecasting benchmarks fall into two camps with complementary weaknesses. Live benchmarks pose questions about events that have not yet occurred, typically drawn from prediction markets or forecasting platforms~\citep{karger2024forecastbench,zeng2025futurex,yang2025prophet}. Since no information about the outcome exists anywhere, they cannot leak outcomes; but feedback arrives only when the event resolves, often weeks or months after the forecast is made~\citep{paleka2026pitfalls}. Every evaluation cycle is gated by the calendar, the question set is dictated by whatever happens to be pending, and scores from different periods cannot be compared because the questions differ~\citep{kucinskas2025forecastbench}. Fixed-corpus benchmarks take the opposite approach: they select events that have already resolved, assemble a corpus of documents published before each question's forecast date, and ask the agent to forecast from that corpus alone~\citep{zou2022autocast,wildman2025btf,liptay2026btf2}. Feedback is immediate and evaluation is reproducible. A main problem, however, is that the agent can research only what the benchmark's creators chose to collect, so evidence outside that corpus stays out of reach however relevant it is. And because the question set is fixed, such benchmarks are static: they saturate~\citep{kiela2021dynabench}, and their outcomes eventually enter model training data~\citep{white2025livebench,paleka2026pitfalls}.

\begin{figure*}[t]
  \centering
  \includegraphics[width=\textwidth]{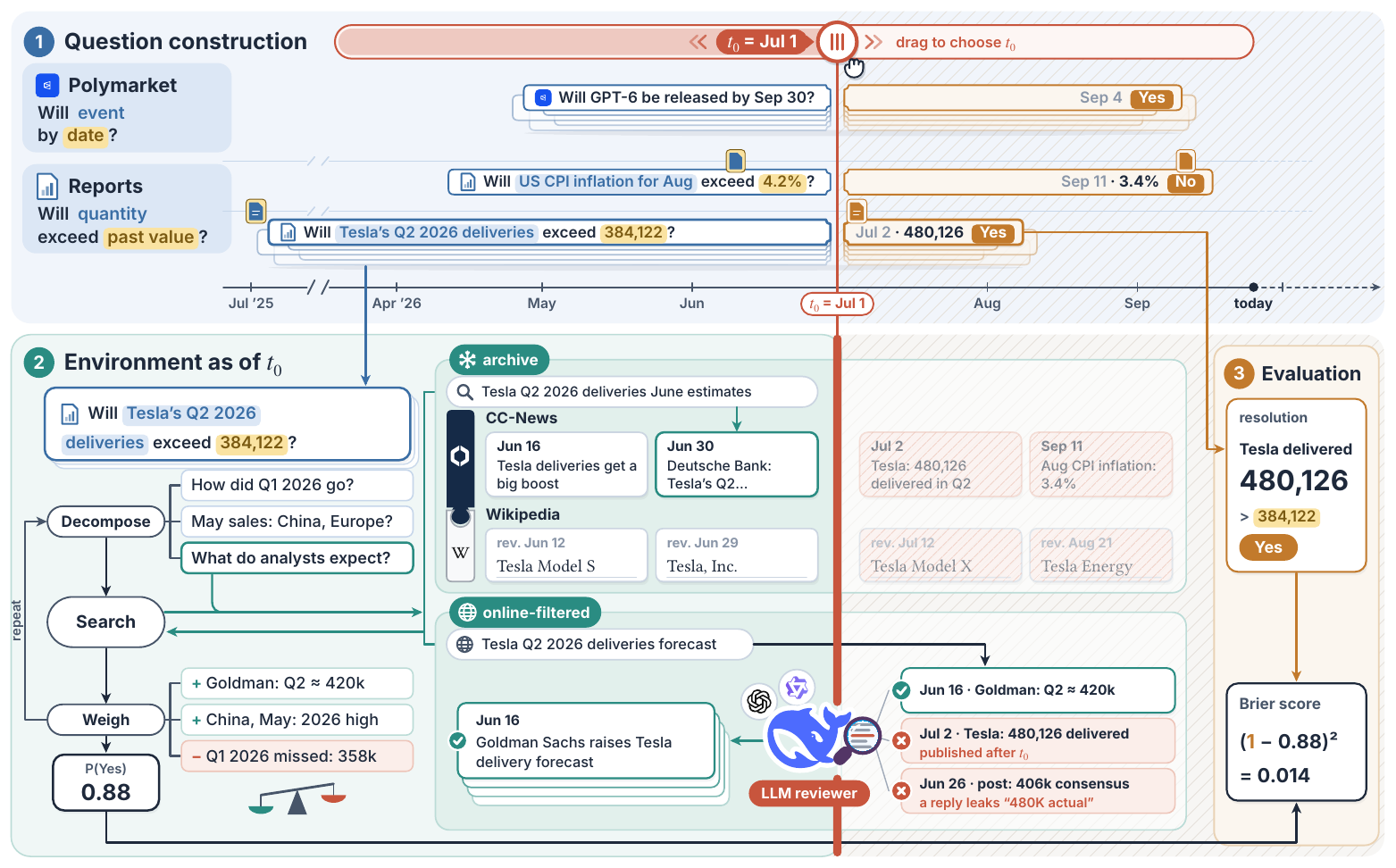}
  \caption{Overview of \ourbenchmarkname.
  The upper part generates questions, and the lower part is the archive environment in which an agent researches a question and gives a probability, scored against the outcome at the bottom right.
  The red line marks the forecast date $t_0$: content to its left was published before $t_0$, and content to its right after $t_0$.}
  \label{fig:overview}
\end{figure*}

We introduce \ourbenchmarkname, a scalable backtesting environment for forecasting agents that removes this trade-off. Our approach treats forecasting as backtesting: questions are posed at historical time points whose outcomes are now known, so feedback is immediate and verifiable, while the agent searches an archive and the live web, both limited to what was published before that time point, so the research itself remains open-ended. \ourbenchmarkname{} has two components (\cref{fig:overview}): a workflow that generates questions, and an archive environment in which agents research them.

\ourbenchmarkname{} currently contains 287 forecasting questions across nine domains: finance, economy, technology, health, sports, culture, geopolitics, weather and politics. Rather than curating them by hand, we build a workflow that discovers candidate events from recurring data releases, news and prediction markets, writes well-specified questions with unambiguous resolution criteria, and resolves each question against its recorded outcome. The workflow takes the forecast date as input, so it can be re-run for new dates and extended to new domains, and the benchmark grows on demand and can be refreshed to stay ahead of model training cutoffs. All current questions are posed at July~1, 2026, after the pretraining cutoffs of all the models we evaluate (\cref{app:evaluation}). %

We use \ourbenchmarkname{} to evaluate agents built on eight frontier models. Retrieval raises the Brier Index of every model, by 3.4 points on average, and brings the leading agents on par with the market consensus, a strong baseline~\citep{wolfers2004prediction,alur2025aia}, which every model trails without retrieval. Yet on about a quarter of the questions, agents still do worse on average than a constant 0.5 forecast, with or without retrieval. Because the environment records every step of the agent's research alongside time-stamped evidence, we can trace these failures to their sources. We identify three error modes: favoring unofficial reports over newer official statements, following a public consensus that proves wrong, and applying correct evidence to the wrong case.

\section{\ourbenchmarkname}
\label{sec:benchmark}

\subsection{Forecasting Task and Evaluation}
\label{sec:task}

We are interested in forecasting consequential real-world events.
To score forecasts without waiting, we pose each question at a past forecast date $t_0$ about an event that has since resolved.
Formally, a question is a yes-or-no statement $q$ with a resolution rule $r$, which names the source that determines the outcome $y\in\{0,1\}$.
For example, a question posed on July~1, 2026, asks whether the first estimate of UK GDP growth for 2026~Q2 will exceed the 0.6\% first published for Q1; it resolves from the Office for National Statistics release of August~13.

Given a question, a forecasting agent returns the probability $p\in[0,1]$ that $y=1$.
To forecast as someone at $t_0$ would, the agent may use only $\mathcal{I}(t_0)$, the information published before $t_0$.
The archive environment of \cref{sec:environment} provides this information.
What counts is when information is published, not when events occur: in the example, the quarter has ended by $t_0$, but its GDP estimate is not yet public.
Besides retrieval, a model may carry later information from training, which no environment can remove; $t_0$ must therefore follow the training cutoffs of the evaluated models.

\label{sec:scoring}%
We score each forecast $p$ against the outcome $y$ with the Brier score~\citep{brier1950verification},
\begin{equation}
  \operatorname{BS}=\frac{1}{N}\sum_{i=1}^{N}(p_i-y_i)^2,
  \label{eq:brier}
\end{equation}
where $N$ is the number of questions with a valid forecast.
It measures accuracy and is strictly proper: a forecaster minimizes its expected score only by reporting its actual belief~\citep{gneiting2007strictly}.
For readability, we report it as the Brier Index~\citep{kucinskas2026brierindex},
\begin{equation}
  \operatorname{BI}=100\,\bigl(1-\sqrt{\operatorname{BS}}\bigr),
  \label{eq:bi}
\end{equation}
on which a constant forecast of 0.5 scores 50 and a perfect forecaster scores 100.

A good Brier score, however, does not show whether the probabilities can be taken at face value.
A forecaster that separates Yes from No events well can score well while being overconfident, yet a decision that weighs a probability against costs needs the probability itself to be right.
To measure this calibration, we also report the expected calibration error~\citep{guo2017calibration},
\begin{equation}
  \operatorname{ECE}=\sum_{b=1}^{10}\frac{n_b}{N}\,\bigl|\bar{p}_b-\bar{y}_b\bigr|,
  \label{eq:ece}
\end{equation}
where ten equal-width probability bins group the forecasts, and bin $b$ holds $n_b$ of them with mean $\bar{p}_b$ and Yes rate $\bar{y}_b$; lower is better.

\subsection{Question Generation}
\label{sec:construction}

To evaluate agents on events that matter, and to renew the questions as models advance, we generate questions with an automatic workflow that takes $t_0$ as input (panel~1 of \cref{fig:overview}).
Renewal is necessary because a set expires once models are trained past its $t_0$.
The workflow draws events from two complementary sources and, in each, keeps those that many people follow.

\paragraph{Recurring releases.}
The first source covers releases such as the earnings of S\&P~100 companies, the decisions of major central banks and the official statistics of major economies.
They meet both needs: many people follow them, and their published schedules and fixed definitions allow questions to be posed at any $t_0$.
For a given $t_0$, an LLM agent selects releases due after $t_0$ from calendars published before it.
A template then turns each release into a question, with a threshold computed by a fixed rule from values published before $t_0$, as in the GDP example; the fixed rule keeps the threshold independent of the outcome.

\paragraph{Emerging events.}
Many consequential events, however, happen only once, such as elections, drug approvals and product launches, and they form the second source.
To find them, the agent searches news and event calendars published before $t_0$ for pending events.
Prediction markets also show which events people follow, so we draw on Polymarket as well.
We search its markets on consequential topics and turn those open at $t_0$ into questions.
Each such question keeps the market's wording and resolution rule, as in ``Will GPT-6 be released by September~30, 2026?''
The market consensus at $t_0$, its price just before $t_0$, is then a forecast of the same question, which gives a human baseline (\cref{sec:exp-main}).

\paragraph{Validation and resolution.}
To ensure validity, a second agent checks that each question the agents write resolves unambiguously, was still open at $t_0$ and reveals nothing published after $t_0$.
Once the resolution date passes, each outcome is recorded from the named source with its evidence and kept out of the agent's input.
Questions whose outcome cannot be established are dropped.

\paragraph{First release.}
Our first release poses 287 questions at $t_0$ = July~1, 2026, 254 written by agents and 33 taken from Polymarket, in nine domains: finance, economy, politics, technology, health, sports, culture, geopolitics and weather (\cref{tab:pipeline,tab:domain-stats}).
Building it also involved manual steps, and some questions were written with outcomes visible; \cref{app:benchmark} lists both.
The workflow is not tied to this date: an earlier run at March~1, 2026, produced 232 agent-written and 300 Polymarket questions (\cref{app:dates}).

\subsection{Comparison with Existing Benchmarks}
\label{sec:bench-compare}

\begin{table}[!ht]
  \centering
  \footnotesize
  \caption{Comparison with representative forecasting benchmarks.
  \cmark{} yes; \pmark{} partly; \xmark{} no.}
  \label{tab:bench-compare}
  \setlength{\tabcolsep}{3pt}
  \begin{tabular}{@{}ll*{5}{c}@{}}
    \toprule
    & Benchmark
      & \colhead{Instant\\scoring}
      & \colhead{Outcome-\\blind}
      & \colhead{Open\\research}
      & \colhead{Answer-\\blind}
      & Calibration \\
    \midrule
    \multirow{3}{*}{Live} & ForecastBench~\citep{karger2024forecastbench} & \xmark & \cmark & \pmark & \cmark & \pmark \\
    & Prophet Arena~\citep{yang2025prophet} & \xmark & \cmark & \pmark & \cmark & \cmark \\
    & FutureX~\citep{zeng2025futurex} & \xmark & \cmark & \cmark & \cmark & \xmark \\
    \midrule
    \multirow{4}{*}{Backtest} & Autocast~\citep{zou2022autocast} & \cmark & \cmark & \xmark & \pmark & \pmark \\
    & Daily Oracle~\citep{dai2024dailyoracle} & \cmark & \xmark & \xmark & \pmark & \xmark \\
    & BTF-2~\citep{liptay2026btf2} & \cmark & \cmark & \pmark & \pmark & \cmark \\
    & ATM~\citep{chai2026atm} & \cmark & \cmark & \cmark & \xmark & \xmark \\
    \midrule
    & \textbf{\ourbenchmarkname} & \cmark & \cmark & \cmark & \cmark & \cmark \\
    \bottomrule
  \end{tabular}
\end{table}

\Cref{tab:bench-compare} compares \ourbenchmarkname{} with representative benchmarks; \cref{app:benchmark} notes where our first release departs from the properties marked there, and \cref{app:related-work} reviews further work.

Live benchmarks, such as ForecastBench, Prophet Arena and FutureX, cannot leak outcomes, but every score waits for events to resolve, and systems evaluated at different times answer different questions.
Other backtests score at once, but they either choose events after the fact, as Daily Oracle does by generating questions from articles that report the outcome, or confine forecasting to a pre-collected corpus, as Autocast does.

BTF-2 is closest to ours in how it selects and scores questions.
Its corpus, however, is crawled for each question while the question is still open, so a set for a new period must be planned and collected before that period, and, as its authors note, models trained later need such a new set.
\ourbenchmarkname{} instead builds a set after the fact for any past $t_0$ that its news archive covers, because the news archive is collected independently of the questions and web pages are screened against $t_0$.
A new model can therefore be evaluated once events after its training cutoff have resolved, without a collection planned in advance.

Agentic Time Machine (ATM) also lets agents research the live web, but its filter reads each question and its known answer, so what the agent sees can depend on the outcome it must forecast.
\ourbenchmarkname{} screens each page by its publication time alone, so the evidence cannot be tilted toward or away from the outcome (\cref{sec:env-web}).
ATM also reuses existing questions and scores point predictions, whereas \ourbenchmarkname{} generates questions for a chosen $t_0$ and scores probabilities, which also reveals calibration.

\section{Archive Environment}
\label{sec:environment}

To let the agent research a question as a forecaster at $t_0$ could, using only information published before $t_0$ (\cref{sec:task}), we build an archive environment (panel~2 of \cref{fig:overview}).
In this environment, the agent decides what to search for and which documents to read, and the environment decides what its search and read tools return.
These tools should deliver as much relevant information as possible while leaking nothing published after $t_0$.
An offline archive of pages captured before $t_0$ meets the second requirement by construction but holds only what was collected.
Live web search reaches far more sources, but its pages must be screened one by one.
We therefore combine the two, so that one search draws on the archive's dated news and reference text and on pages that only the web can reach.

For a request of $N$ results, the archive supplies its top $\lceil N/2\rceil$ documents, and the web supplies the pages that pass screening among its top $\lfloor N/2\rfloor$ candidates; withheld pages are not replaced.
The fixed shares keep archive documents, which cannot leak, in every result list, so the agent never depends on screened web pages alone.
A URL found in both keeps only its archive copy, whose capture date needs no review.
Either source can also be searched alone, which \cref{sec:exp-retrieval} uses to isolate what each contributes.

\subsection{Offline archive}
\label{sec:env-archive}

\paragraph{Contents.}
To forecast well from an archive, a forecaster needs two kinds of information: background on the entities and topics of a question, and the latest developments before $t_0$.
For background, we use Wikipedia.
Because it stores every revision with its time, we take the last revision before $t_0$ of the articles on the entities and topics in the questions (\cref{app:environment}).
For recent developments, we use CC-News, Common Crawl's continuous crawl of news sites~\citep{nagel2016ccnews}, and keep the English pages captured before $t_0$.
A page captured before $t_0$ was public by then, so the archive needs no page-by-page review and cannot leak through a reviewer's mistakes.
For $t_0$ = July~1, 2026, it holds about 20.1 million news captures and 684 Wikipedia revisions.

\paragraph{Retrieval.}
When the agent searches the archive, BM25 retrieves candidates from both parts, and a rule-based scorer removes duplicates and reranks them.
All term statistics come from documents before $t_0$, so no signal from after $t_0$ shapes the ranking.
BTF-2 also gives agents an offline corpus, but its RetroSearch orders results by live Google search before dropping pages outside the corpus~\citep{liptay2026btf2,wildman2025btf}.
Pages that became prominent because of later events can then rise to the top, a possible leak that its developers acknowledge, and the same query can return different results in different runs.
Our ranking uses the archive alone, so the same query returns the same results in every run.
A search result shows the beginning of each document, and the read tool returns its full text (\cref{app:environment}).

\subsection{Screened web search}
\label{sec:env-web}

The archive, however, misses sites and pages that CC-News did not capture.
To reach them, the search tool also asks a live search engine for results dated before $t_0$.

\paragraph{Search and cleaning.}
The search engine's date limit only narrows the candidates, because an old URL may now show updated content.
We therefore fetch each web page before the agent sees it and judge it by the text it shows now.
To keep market prices out of the agent's view, we reject pages on known prediction-market sites.
From the other pages, we strip navigation and advertisements, and we replace Wikipedia pages with their last revision before $t_0$, which needs no review.

\paragraph{Temporal review.}
Every other page goes to an LLM reviewer that sees neither the question nor its answer, so that its decision depends on timing alone (\cref{app:environment}).
The reviewer first removes leftover blocks, such as recommendation lists and live widgets, whose newer headlines would otherwise count against an old article.
It then judges whether the remaining text was public before $t_0$.
What counts is publication, not when events occur, so a forecast or plan published before $t_0$ passes.
When timing is unclear, the reviewer can search the web to check, and what it finds never reaches the agent.
The filter of Agentic Time Machine, by contrast, sees the answer, so it can withhold a legitimate forecast for matching the outcome, as it did with an award-nomination forecast published two days before the announcement~\citep{chai2026atm}.
Our reviewer passes such forecasts, which a forecaster at $t_0$ would have weighed.

\paragraph{Delivery.}
Based on this judgment, a page is withheld entirely if it contains information from after $t_0$ or its timing stays unclear.
Agentic Time Machine passes pages of unclear timing, and in one of its cases an undated post quoting a price from another day misled the agent.
The cost of our choice is that some legitimate pages of unclear timing are lost.
Withholding the whole page, rather than cutting out the later passages, avoids passing text whose remaining context could still reveal later events.
For a page that passes, the read tool returns the cleaned text, and a search result shows an excerpt of it.
The search engine's own titles, snippets and dates never reach the agent, because they are generated today and can reflect later events.
Every page is logged with its decision, so that decisions can be audited; \cref{app:environment} gives the reviewer's instructions.
In the \texttt{online-filtered} runs, the reviewer withheld 16.0\% of 302{,}561 pages.
Reviewing 3{,}087 pages at both our low reasoning effort and high effort gave the same decision for 95.7\% of them; only 14 (0.5\%) passed at low effort yet were judged at high effort to disclose later information.

\section{Experiments}
\label{sec:experiments}

\providecommand{\expfigdir}{paras/figures/experiments}
\providecommand{\expfig}[2]{%
  \IfFileExists{\expfigdir/#2}%
    {\includegraphics[width=#1]{\expfigdir/#2}}%
    {\fbox{\parbox[c][0.2\textheight][c]{\dimexpr#1-2\fboxsep-2\fboxrule\relax}%
      {\centering\ttfamily missing figure: \expfigdir/#2}}}}

\subsection{Experimental setup}
\label{sec:exp-setup}

\paragraph{Environment and data.}
We evaluate every model in four information settings: \texttt{none}, without tools; \texttt{archive} and \texttt{online-filtered}, with only the archive or only screened web search; and \texttt{full}, our standard setting, with both (\cref{sec:environment}).
All 287 questions come from the release of \cref{sec:construction}, posed at $t_0$ = July~1, 2026, and every model is asked all of them in every setting.\footnote{A few runs lack forecasts for some questions, mostly political questions that GPT-6 Astra and GPT-6 Sol declined; each run is scored on the questions it answered (\cref{tab:usage}).}

\paragraph{Compared models.}
We compare \textbf{(1)} three proprietary models: GPT-6 Astra, GPT-6 Sol and Claude Opus~5.5; \textbf{(2)} five open-weight models: DeepSeek V4.1 Flash (DS V4.1 Flash), Kimi~K3, GLM~5.3, Qwen3.8-27B and Qwen3.6-35B-A3B; and \textbf{(3)} two models trained to return calibrated probabilities, JEV~1.13.0~\citep{almeida2026jev} and StartLux-Decision 27B AutoThink (StartLux~27B;~\citealp{zhuang2026startlux}), which answer without retrieval.
As references, we use the constant forecast of 0.5 and the market consensus on Polymarket (\cref{sec:exp-main}).
Most models were released after $t_0$, but without retrieval the best of them scores 55.0~BI, only 5 above the constant forecast, so none appears to know the outcomes from training (\cref{tab:models}).

\paragraph{Implementation and metrics.}
Every model receives the same prompt, searches the same sources and makes at most 100 tool calls per question, and forecasts each question once per setting; ChatGPT models run in the Codex CLI and Claude in Claude Code with their built-in search replaced by our tools, and the other models in our agent loop (\cref{app:environment,app:evaluation}).
We measure accuracy with the BI and calibration with the ECE (\cref{sec:task}).

\subsection{Comparison among models and with prediction markets}
\label{sec:exp-main}
\label{sec:exp-overall}

\begin{figure}[t]
  \centering
  \expfig{0.545\linewidth}{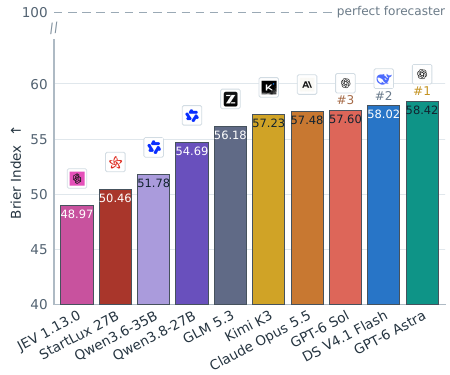}\hfill
  \expfig{0.445\linewidth}{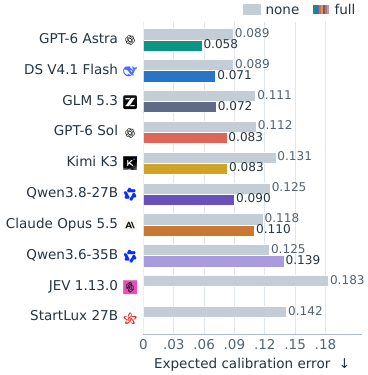}
  \caption{BI under \texttt{full} (left) and ECE under \texttt{none} and \texttt{full} (right) of each model.
  Lower ECE is better; JEV~1.13.0 and StartLux~27B do not retrieve, so they appear with their \texttt{none} scores.}
  \label{fig:models}
\end{figure}

\paragraph{Comparison among models.}
To compare the models in the standard setting, \cref{fig:models} shows their BI and ECE under \texttt{full}.
Five models, three proprietary and two open-weight, form a leading group at 57 to 58~BI, while the weakest model and the two models that do not retrieve score near the constant forecast.
Even the best model removes only about 31\% of the squared error of the constant forecast.

The ranking depends less on what a model knows than on how much it researches.
Accuracy without tools is unrelated to the gain from retrieval (Spearman $\rho=0.24$ across the eight models), whereas the number of pages a model reads per question tracks it closely ($\rho=0.83$; \cref{tab:retrieval,tab:usage}).
Kimi~K3 and DeepSeek V4.1 Flash read the most pages and gain the most, which lifts them into the leading group, whereas Claude Opus~5.5, the most accurate model without tools, reads less than one page per question and gains only 2.4~BI.

Calibration does not follow accuracy: Claude Opus~5.5 scores within 1~BI of GPT-6 Astra, yet its ECE is 0.110 against 0.058.
Retrieval lowers ECE for 7 of the 8 models, but not for Qwen3.6-35B, whose ECE rises from 0.125 to 0.139 although its BI improves.

\paragraph{Comparison with prediction markets.}
To compare the models with human forecasters, we use the 54 questions linked to Polymarket: 33 taken from Polymarket and 21 generated questions that match a contract.
The market consensus, the market's price about one minute before $t_0$, is a forecast made at the same cutoff as the models' and a strong baseline: prior work has found it hard to beat~\citep{karger2024forecastbench,alur2025aia,mostafa2026timeseek}, turning forecasts into trading gains is harder still~\citep{wang2026beyond}, and roughly 69\% of Polymarket users lose money~\citep{akey2026polymarket}, which suggests that forecasts at the market's level may already outperform most participants.

\Cref{fig:market} compares each model with the market in the four settings.
Under \texttt{full}, the leading agents perform at about the market's level: six models lie between 1.8~BI below and 2.7~BI above the market.
Without retrieval, every model falls below the market, by 1.8 to 10.8~BI.
The market price aggregates what traders learned before $t_0$, and retrieval lets an agent gather the same public information itself, which plausibly explains why the gap closes only with retrieval.
Prior work has instead found LLM forecasters trailing liquid markets~\citep{alur2025aia} or beating a market only shortly after it opens~\citep{mostafa2026timeseek}.

\afterpage{%
\begin{figure}[t]
  \centering
  \expfig{\linewidth}{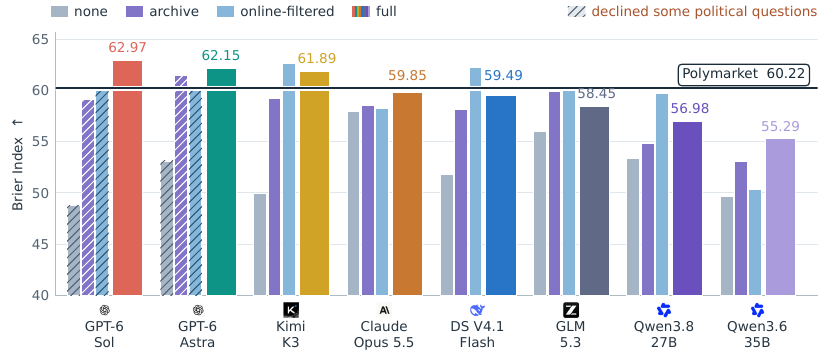}
  \caption{BI of each model on the 54 questions linked to Polymarket.
  The line marks the market consensus at $t_0$, numbers give the BI under \texttt{full}, and hatched bars omit declined political questions.}
  \label{fig:market}
\end{figure}
}

\subsection{Comparison of retrieval settings}
\label{sec:exp-retrieval}

\begin{table}[t]
  \centering
  \caption{BI of each model in the four information settings, each on the questions the model answered in that setting, with the difference from \texttt{none} in parentheses.
  Bold marks the highest and underlining the second highest score and gain in each column.}
  \label{tab:retrieval}
  \footnotesize
  \setlength{\tabcolsep}{5pt}
  \begin{tabular}{lcccc}
    \toprule
    Model & \texttt{none} & \texttt{archive} & \texttt{online-filtered} & \texttt{full} \\
    \midrule
    GPT-6 Astra & \underline{54.33} & \textbf{57.57} (+3.24) & \underline{58.33} (+4.00) & \textbf{58.42} (+4.09) \\
    DeepSeek V4.1 Flash & 53.78 & \underline{57.24} (\underline{+3.46}) & \textbf{58.48} (\underline{+4.70}) & \underline{58.02} (\underline{+4.24}) \\
    GPT-6 Sol & 53.51 & 56.48 (+2.97) & 57.59 (+4.08) & 57.60 (+4.09) \\
    Claude Opus~5.5 & \textbf{55.04} & 56.48 (+1.44) & 57.42 (+2.38) & 57.48 (+2.44) \\
    Kimi~K3 & 51.39 & 56.36 (\textbf{+4.97}) & 57.81 (\textbf{+6.42}) & 57.23 (\textbf{+5.84}) \\
    GLM~5.3 & 53.15 & 55.77 (+2.62) & 56.35 (+3.20) & 56.18 (+3.03) \\
    Qwen3.8-27B & 52.36 & 54.88 (+2.52) & 55.20 (+2.84) & 54.69 (+2.33) \\
    Qwen3.6-35B-A3B & 50.31 & 50.48 (+0.17) & 50.92 (+0.61) & 51.78 (+1.47) \\
    \bottomrule
  \end{tabular}
\end{table}

To test whether each source adds useful information, and whether combining them adds more, \cref{tab:retrieval} compares the four settings for each model.
Retrieval improves every model in every setting.
The archive alone, which cannot leak by construction, already yields about four fifths of the gain of \texttt{full}, so most of the improvement reflects information beyond what the models learned in training rather than leaked outcomes.

The three retrieval settings, however, lie close together.
\texttt{online-filtered} scores slightly higher than \texttt{archive} for all eight models, but for no model does \texttt{full} differ from either single source with a 95\% interval that excludes zero (paired bootstrap over questions; \cref{app:settings}).
This pattern suggests that screened web search adds little information from after $t_0$ beyond the archive, which predates $t_0$ by construction.
It also suggests that, on average, neither source lacks much of what the other provides, although the archive falls short in three domains (\cref{sec:exp-domains}).

\subsection{Performance across domains}
\label{sec:exp-domains}

\afterpage{%
\begin{figure}[t]
  \centering
  \expfig{0.85\linewidth}{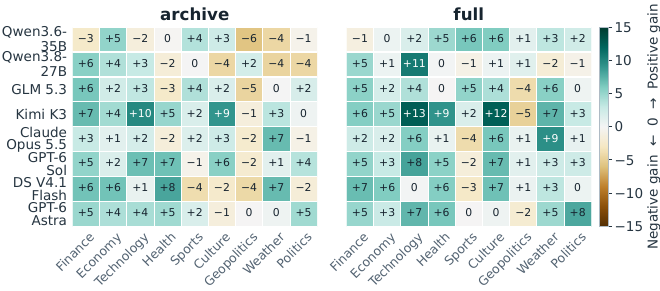}
  \caption{Gain in BI from retrieval, by model and domain.
  Gain over \texttt{none} under \texttt{archive} and \texttt{full}, rounded to whole BI; domains contain 13 to 77 questions.}
  \label{fig:domains}
\end{figure}
}

Forecasting performance differs widely across domains: under \texttt{full}, the BI averaged over models ranges from 49.1 on Culture and 49.8 on Weather to 63.2 on Health.
Retrieval, too, helps unevenly, and \cref{fig:domains} breaks its gains down by domain under \texttt{archive}, which searches the archive alone, and under \texttt{full}, which adds screened web search.
Overall, retrieval raises the BI in most model--domain pairs: 48 of 72 under \texttt{archive} and 60 of 72 under \texttt{full}.
Adding web search matters most in Technology, Health and Culture, where the gain averaged over models rises from about 1.7 to 3.5~BI under \texttt{archive} to about 4.0 to 6.3~BI under \texttt{full}.
Finance and Economy gain about 2.7 to 4.5~BI in either setting.

One explanation is that questions in these three domains often ask whether an announced event, such as a model release, a drug approval or a film opening, happens as planned.
The decisive sources are then company, regulator and industry pages, which web search reaches but a news archive may not contain (\cref{sec:env-web}).
News, by contrast, closely covers the scheduled releases behind finance and economy questions.

Retrieval can also lower accuracy.
In both settings, every model loses BI in at least one domain, by up to 5.6~BI under \texttt{archive} (Qwen3.6-35B on Geopolitics) and 4.8~BI under \texttt{full} (Kimi~K3 on Geopolitics).
Geopolitics gains little or loses on average, plausibly because its coverage before $t_0$ consists largely of commentary and competing predictions about open developments.
Forecasting performance should therefore be reported together with the domains and the setting in which it was measured.

\subsection{Failure analysis}
\label{sec:exp-failures}

To find failures that aggregate scores hide, we classify each question by whether its Brier score, averaged over models, exceeds 0.25, the score of the constant forecast, without retrieval and with \texttt{full}.
Retrieval corrects 41 questions and misleads 11, while 76, about a quarter of the benchmark, stay above 0.25 in both settings; on 28 of them, no model beats the constant forecast.
Traces of representative questions show why (\cref{app:failures}).

\paragraph{Retrieval helps when decisive evidence is already public.}
In the traced corrections, the decisive evidence, such as a central bank's split vote and guidance, was published before $t_0$ and replaced the trend that models extrapolate without tools.

\paragraph{Retrieval misleads when informal signals meet official guidance.}
Several models let a concrete but unofficial report, such as a launch window relayed on social media, outweigh a newer official statement, which suggests that they weigh how specific a signal is more than who issued it and when.
Adding the archive reduces this error in the traced case; most misled questions, 7 of 11, concern finance.

\paragraph{Shared errors come from the public record itself.}
When the consensus at $t_0$ is wrong, as with box-office forecasts that all overestimated a film's opening, retrieval returns that consensus and cannot correct it; models treat the few dissenting forecasts as tail risk.

\paragraph{Evidence is applied to the wrong case.}
In the traced cases, models match evidence to a question by topic rather than by the exact event, so an authoritative fact about one case, such as the production timeline of one coin, is applied to another.

\section{Conclusion}
\label{sec:conclusion}

We introduced \ourbenchmarkname{}, a scalable backtesting environment for forecasting agents that combines automated question generation with open-ended research under historical information constraints. By pairing resolved events with an offline archive and temporally screened web search, it provides immediate, verifiable feedback without restricting agents to a pre-collected corpus. Our evaluation of latest models on 287 questions across nine domains shows that retrieval generally improves forecasting accuracy and calibration, but its benefits vary across models and domains. Several agents score on par with the market consensus on the questions linked to Polymarket contracts, while our case studies reveal failures in how retrieved evidence is weighed and applied. These findings motivate improving not only information access, but also the reasoning that turns evidence into probabilistic judgments. With renewable questions, open-ended research, and immediate outcome feedback, \ourbenchmarkname{} provides a foundation for faster evaluation and future outcome-based training of forecasting agents.

\clearpage
\section*{AI Use Statement}

We used generative AI tools to assist with drafting and editing the
manuscript and creating figures. LLMs were also used for event
discovery, benchmark question construction, selection, and validation,
as well as cleaning and temporal screening of retrieved web content.
The authors reviewed and revised the AI-assisted text and figures
for accuracy and consistency with the methods and results.
The authors take responsibility for the final content of this work,
including all AI-assisted text, claims, and research artifacts.

\section*{Ethics Statement}

This work develops a benchmark for evaluating forecasting agents
using questions and evidence drawn from publicly available sources.
The selection of events and reliance on English-language news may
introduce coverage biases. Forecasts can be inaccurate or
miscalibrated, and performance on this benchmark should not be
interpreted as evidence that a system is suitable for high-stakes
deployment. To support research integrity, we document temporal
screening procedures and trace audits, including their limitations (\cref{app:limitations}).
Responsible use and redistribution of the underlying sources
require attention to attribution, licensing, and privacy.

\section*{Reproducibility Statement}

We will release every question set of \ourbenchmarkname{} on Hugging Face, starting with the first release and the set posed at March~1, 2026, together with their resolution keys and the materials needed to rebuild the retrieval archive: the manifests of the CC-News captures and the identifiers of the Wikipedia revisions.
The code is released with the data: a few commands build the archive environment and evaluate an agent in any of the four information settings, with the prompts, tool definitions and reviewer instructions of \cref{app:environment,app:evaluation}.
New question sets built with the same pipeline will feed a live leaderboard, so that agents can be compared on questions that postdate their training.
The full research traces of the reported runs, over 100~GB, will be shared with teams that request them.

\bibliography{reference}

\begin{thebibliography}{41}
\providecommand{\natexlab}[1]{#1}
\providecommand{\url}[1]{\texttt{#1}}
\expandafter\ifx\csname urlstyle\endcsname\relax
  \providecommand{\doi}[1]{doi: #1}\else
  \providecommand{\doi}{doi: \begingroup \urlstyle{rm}\Url}\fi

\bibitem[Akey et~al.(2026)Akey, Gr{\'e}goire, Harvie, and
  Martineau]{akey2026polymarket}
Pat Akey, Vincent Gr{\'e}goire, Nicolas Harvie, and Charles Martineau.
\newblock {Who Wins and Who Loses in Prediction Markets? Evidence from
  Polymarket}.
\newblock CEPR Discussion Paper DP21615, Centre for Economic Policy Research,
  2026.
\newblock URL \url{https://cepr.org/publications/dp21615}.

\bibitem[Almeida(2026)]{almeida2026jev}
Diogo Almeida.
\newblock Introducing system one models \& {Jev}.
\newblock TypeSafe AI blog, 2026.
\newblock URL
  \url{https://typesafe.ai/blog/introducing-system-one-models-and-jev}.

\bibitem[Alur et~al.(2025)Alur, Stadie, Kang, Chen, McManus, Rickert, Lee,
  Federici, Zhu, Fogerty, Williamson, Lozinski, Linsky, and
  Sekhon]{alur2025aia}
Rohan Alur, Bradly~C. Stadie, Daniel Kang, Ryan Chen, Matt McManus, Michael
  Rickert, Tyler Lee, Michael Federici, Richard Zhu, Dennis Fogerty, Hayley
  Williamson, Nina Lozinski, Aaron Linsky, and Jasjeet~S. Sekhon.
\newblock {AIA Forecaster: Technical Report}, 2025.
\newblock URL \url{https://arxiv.org/abs/2511.07678}.
\newblock arXiv:2511.07678.

\bibitem[Brier(1950)]{brier1950verification}
Glenn~W. Brier.
\newblock Verification of forecasts expressed in terms of probability.
\newblock \emph{Monthly Weather Review}, 78\penalty0 (1):\penalty0 1--3, 1950.

\bibitem[Chai et~al.(2026)Chai, Zheng, Liu, Lu, Zhou, Wang, Zhang, and
  Chen]{chai2026atm}
Jingyi Chai, Bingyang Zheng, Xiangrui Liu, Hao Lu, Zihang Zhou, Tianchen Wang,
  Kemeng Zhang, and Siheng Chen.
\newblock {Agentic Time Machine as an Infrastructure for Future-Event
  Forecasting}, 2026.
\newblock URL \url{https://arxiv.org/abs/2606.21013}.
\newblock arXiv:2606.21013.

\bibitem[Chandak et~al.(2025)Chandak, Goel, Prabhu, Hardt, and
  Geiping]{chandak2025openforesight}
Nikhil Chandak, Shashwat Goel, Ameya Prabhu, Moritz Hardt, and Jonas Geiping.
\newblock {Scaling Open-Ended Reasoning to Predict the Future}, 2025.
\newblock URL \url{https://arxiv.org/abs/2512.25070}.
\newblock arXiv:2512.25070.

\bibitem[Dai et~al.(2025)Dai, Teehan, and Ren]{dai2024dailyoracle}
Hui Dai, Ryan Teehan, and Mengye Ren.
\newblock Are {LLMs} prescient? {A} continuous evaluation using daily news as
  the oracle.
\newblock In \emph{Proceedings of the 42nd International Conference on Machine
  Learning}, volume 267 of \emph{Proceedings of Machine Learning Research},
  pp.\  12179--12213, 2025.
\newblock URL \url{https://proceedings.mlr.press/v267/dai25l.html}.

\bibitem[Gneiting \& Raftery(2007)Gneiting and Raftery]{gneiting2007strictly}
Tilmann Gneiting and Adrian~E. Raftery.
\newblock Strictly proper scoring rules, prediction, and estimation.
\newblock \emph{Journal of the American Statistical Association}, 102\penalty0
  (477):\penalty0 359--378, 2007.

\bibitem[Goel et~al.(2026)Goel, Chandak, Arun, Prabhu, Staab, Hardt,
  Andriushchenko, and Geiping]{goel2026futuresim}
Shashwat Goel, Nikhil Chandak, Arvindh Arun, Ameya Prabhu, Steffen Staab,
  Moritz Hardt, Maksym Andriushchenko, and Jonas Geiping.
\newblock {FutureSim: Replaying World Events to Evaluate Adaptive Agents},
  2026.
\newblock URL \url{https://arxiv.org/abs/2605.15188}.
\newblock arXiv:2605.15188.

\bibitem[Gu et~al.(2026)Gu, Kagan, Wu, Sun, and Xu]{gu2026prophets}
Anri Gu, Nicole Kagan, Jibang Wu, Alec Sun, and Haifeng Xu.
\newblock When do prophets profit in prediction markets?, 2026.
\newblock URL \url{https://arxiv.org/abs/2607.06166}.
\newblock arXiv:2607.06166.

\bibitem[Guo et~al.(2017)Guo, Pleiss, Sun, and Weinberger]{guo2017calibration}
Chuan Guo, Geoff Pleiss, Yu~Sun, and Kilian~Q. Weinberger.
\newblock On calibration of modern neural networks.
\newblock In \emph{Proceedings of the 34th International Conference on Machine
  Learning}, volume~70 of \emph{Proceedings of Machine Learning Research}, pp.\
   1321--1330, 2017.

\bibitem[Halawi et~al.(2024)Halawi, Zhang, Yueh-Han, and
  Steinhardt]{halawi2024approaching}
Danny Halawi, Fred Zhang, Chen Yueh-Han, and Jacob Steinhardt.
\newblock Approaching human-level forecasting with language models.
\newblock In \emph{Advances in Neural Information Processing Systems},
  volume~37, pp.\  50426--50468, 2024.
\newblock \doi{10.52202/079017-1598}.

\bibitem[Han et~al.(2026)Han, Zhang, Wei, Gao, Yue, Chen, Zhuang, Guan, He, Li,
  Duan, Shi, Hu, and Zheng]{han2026futureworld}
Zhixin Han, Yanzhi Zhang, Chuyang Wei, Maohang Gao, Xiawei Yue, Kefei Chen,
  Yu~Zhuang, Haoxiang Guan, Jiyan He, Jian Li, Yitong Duan, Yu~Shi, Mengting
  Hu, and Shuxin Zheng.
\newblock {FutureWorld: A Live Reinforcement Learning Environment for
  Predictive Agents with Real-World Outcome Rewards}, 2026.
\newblock URL \url{https://arxiv.org/abs/2604.26733}.
\newblock arXiv:2604.26733.

\bibitem[Jimenez et~al.(2024)Jimenez, Yang, Wettig, Yao, Pei, Press, and
  Narasimhan]{jimenez2024swebench}
Carlos~E. Jimenez, John Yang, Alexander Wettig, Shunyu Yao, Kexin Pei, Ofir
  Press, and Karthik Narasimhan.
\newblock {SWE-bench}: Can language models resolve real-world {GitHub} issues?
\newblock In \emph{International Conference on Learning Representations}, pp.\
  54107--54157, 2024.
\newblock URL
  \url{https://proceedings.iclr.cc/paper_files/paper/2024/hash/edac78c3e300629acfe6cbe9ca88fb84-Abstract-Conference.html}.

\bibitem[Karger et~al.(2025)Karger, Bastani, Yueh-Han, Jacobs, Halawi, Zhang,
  and Tetlock]{karger2024forecastbench}
Ezra Karger, Houtan Bastani, Chen Yueh-Han, Zachary Jacobs, Danny Halawi, Fred
  Zhang, and Philip~E. Tetlock.
\newblock {ForecastBench}: A dynamic benchmark of {AI} forecasting
  capabilities.
\newblock In \emph{International Conference on Learning Representations}, pp.\
  93943--93980, 2025.
\newblock URL
  \url{https://proceedings.iclr.cc/paper_files/paper/2025/hash/ea74e45a229dac70b5b63b28d8934db6-Abstract-Conference.html}.

\bibitem[Kiela et~al.(2021)Kiela, Bartolo, Nie, Kaushik, Geiger, Wu, Vidgen,
  Prasad, Singh, Ringshia, Ma, Thrush, Riedel, Waseem, Stenetorp, Jia, Bansal,
  Potts, and Williams]{kiela2021dynabench}
Douwe Kiela, Max Bartolo, Yixin Nie, Divyansh Kaushik, Atticus Geiger,
  Zhengxuan Wu, Bertie Vidgen, Grusha Prasad, Amanpreet Singh, Pratik Ringshia,
  Zhiyi Ma, Tristan Thrush, Sebastian Riedel, Zeerak Waseem, Pontus Stenetorp,
  Robin Jia, Mohit Bansal, Christopher Potts, and Adina Williams.
\newblock Dynabench: Rethinking benchmarking in {NLP}.
\newblock In \emph{Proceedings of the 2021 Conference of the North American
  Chapter of the Association for Computational Linguistics: Human Language
  Technologies}, pp.\  4110--4124, Online, June 2021. Association for
  Computational Linguistics.
\newblock \doi{10.18653/v1/2021.naacl-main.324}.

\bibitem[Kucinskas et~al.(2025)Kucinskas, Bastani, and
  Karger]{kucinskas2025forecastbench}
Simas Kucinskas, Houtan Bastani, and Ezra Karger.
\newblock {ForecastBench}: Updated ranking methodology.
\newblock Technical report, ForecastBench, 2025.
\newblock URL
  \url{https://www.forecastbench.org/assets/pdfs/forecastbench_updated_methodology.pdf}.
\newblock Released 3 October 2025.

\bibitem[Kucinskas et~al.(2026)Kucinskas, Bastani, and
  Reynolds]{kucinskas2026brierindex}
Simas Kucinskas, Houtan Bastani, and Matt Reynolds.
\newblock Making forecasting scores easier to interpret: Introducing the {Brier
  Index}.
\newblock Substack post, 2026.
\newblock URL
  \url{https://forecastingresearch.substack.com/p/introducing-the-brier-index}.

\bibitem[Kwa et~al.(2025)Kwa, West, Becker, Deng, Garcia, Hasin, Jawhar,
  Kinniment, Rush, {Von Arx}, Bloom, Broadley, Du, Goodrich, Jurkovic, Miles,
  Nix, Lin, Painter, Parikh, Rein, {Koba Sato}, Wijk, Ziegler, Barnes, and
  Chan]{kwa2025measuring}
Thomas Kwa, Ben West, Joel Becker, Amy Deng, Katharyn Garcia, Max Hasin, Sami
  Jawhar, Megan Kinniment, Nate Rush, Sydney {Von Arx}, Ryan Bloom, Thomas
  Broadley, Haoxing Du, Brian Goodrich, Nikola Jurkovic, Luke~Harold Miles,
  Seraphina Nix, Tao Lin, Chris Painter, Neev Parikh, David Rein, Lucas~Jun
  {Koba Sato}, Hjalmar Wijk, Daniel~M. Ziegler, Elizabeth Barnes, and Lawrence
  Chan.
\newblock Measuring {AI} ability to complete long software tasks.
\newblock In \emph{Advances in Neural Information Processing Systems},
  volume~38, pp.\  92213--92266, 2025.
\newblock \doi{10.52202/085713-3086}.

\bibitem[Liptay et~al.(2026)Liptay, Schwarz, Poyiadzi, Wildman, and
  Bosse]{liptay2026btf2}
Tom Liptay, Dan Schwarz, Rafael Poyiadzi, Jack Wildman, and Nikos~I. Bosse.
\newblock {Evaluating Strategic Reasoning in Forecasting Agents}, 2026.
\newblock URL \url{https://arxiv.org/abs/2604.26106}.
\newblock arXiv:2604.26106.

\bibitem[Mellers et~al.(2015)Mellers, Stone, Murray, Minster, Rohrbaugh,
  Bishop, Chen, Baker, Hou, Horowitz, Ungar, and
  Tetlock]{mellers2015identifying}
Barbara Mellers, Eric Stone, Terry Murray, Angela Minster, Nick Rohrbaugh,
  Michael Bishop, Eva Chen, Joshua Baker, Yuan Hou, Michael Horowitz, Lyle
  Ungar, and Philip Tetlock.
\newblock Identifying and cultivating superforecasters as a method of improving
  probabilistic predictions.
\newblock \emph{Perspectives on Psychological Science}, 10\penalty0
  (3):\penalty0 267--281, 2015.
\newblock \doi{10.1177/1745691615577794}.

\bibitem[Mostafa et~al.(2026)Mostafa, Shastri, and Lee]{mostafa2026timeseek}
Hamza Mostafa, Om~Shastri, and Dennis Lee.
\newblock {TimeSeek: Temporal Reliability of Agentic Forecasters}, 2026.
\newblock URL \url{https://arxiv.org/abs/2604.04220}.
\newblock arXiv:2604.04220.

\bibitem[Murphy(2026)]{murphy2026blf}
Kevin Murphy.
\newblock {Agentic Forecasting using Sequential Bayesian Updating of Linguistic
  Beliefs}, 2026.
\newblock URL \url{https://arxiv.org/abs/2604.18576}.
\newblock arXiv:2604.18576.

\bibitem[Nagel(2016)]{nagel2016ccnews}
Sebastian Nagel.
\newblock News dataset available.
\newblock Common Crawl blog, 2016.
\newblock URL \url{https://commoncrawl.org/blog/news-dataset-available}.

\bibitem[Paleka et~al.(2026)Paleka, Goel, Geiping, and
  Tram{\`e}r]{paleka2026pitfalls}
Daniel Paleka, Shashwat Goel, Jonas Geiping, and Florian Tram{\`e}r.
\newblock Pitfalls in evaluating language model forecasters.
\newblock In \emph{International Conference on Learning Representations}, pp.\
  114207--114227, 2026.
\newblock URL
  \url{https://proceedings.iclr.cc/paper_files/paper/2026/hash/b9d69c72e4587ddf3fa6352890aefc4f-Abstract-Conference.html}.

\bibitem[Patwardhan et~al.(2026)Patwardhan, Dias, Proehl, Kim, Wang, Watkins,
  Fishman, Aljubeh, Thacker, Fauconnet, Kim, Chao, Miserendino, Chabot, Li,
  Sharman, Barr, Glaese, and Tworek]{patwardhan2026gdpval}
Tejal Patwardhan, Rachel Dias, Elizabeth Proehl, Grace Kim, Michele Wang,
  Olivia Watkins, Sim{\'o}n~Posada Fishman, Marwan Aljubeh, Phoebe Thacker,
  Laurance Fauconnet, Natalie~S. Kim, Patrick Chao, Samuel Miserendino, Gildas
  Chabot, David Li, Michael Sharman, Alexandra Barr, Amelia Glaese, and Jerry
  Tworek.
\newblock {GDPval}: Evaluating {AI} model performance on real-world
  economically valuable tasks.
\newblock In \emph{International Conference on Learning Representations}, pp.\
  24005--24040, 2026.
\newblock URL
  \url{https://proceedings.iclr.cc/paper_files/paper/2026/hash/290c2430f91912204f30bbcc990fff1d-Abstract-Conference.html}.

\bibitem[Schoenegger et~al.(2024)Schoenegger, Tuminauskaite, Park, Bastos, and
  Tetlock]{schoenegger2024silicon}
Philipp Schoenegger, Indre Tuminauskaite, Peter~S. Park, Rafael Valdece~Sousa
  Bastos, and Philip~E. Tetlock.
\newblock Wisdom of the silicon crowd: {LLM} ensemble prediction capabilities
  rival human crowd accuracy.
\newblock \emph{Science Advances}, 10\penalty0 (45):\penalty0 eadp1528, 2024.
\newblock \doi{10.1126/sciadv.adp1528}.

\bibitem[Su et~al.(2026)Su, Xing, Dong, Zhong, Wang, Zhu, Chen, Wang, Deng,
  Zhu, Liu, Li, Yu, Chen, Bing, and Dai]{su2026miroflow}
Shiqian Su, Sen Xing, Xuan Dong, Muyan Zhong, Bin Wang, Xizhou Zhu, Yuntao
  Chen, Wenhai Wang, Yue Deng, Pengxiang Zhu, Ziyuan Liu, Tiantong Li, Jiaheng
  Yu, Zhe Chen, Lidong Bing, and Jifeng Dai.
\newblock {MiroFlow: Towards High-Performance and Robust Open-Source Agent
  Framework for General Deep Research Tasks}, 2026.
\newblock URL \url{https://arxiv.org/abs/2602.22808}.
\newblock arXiv:2602.22808.

\bibitem[Tetlock \& Gardner(2015)Tetlock and
  Gardner]{tetlock2015superforecasting}
Philip~E. Tetlock and Dan Gardner.
\newblock \emph{Superforecasting: The Art and Science of Prediction}.
\newblock Crown Publishers, New York, 2015.
\newblock ISBN 978-0-8041-3669-3.

\bibitem[Turtel et~al.(2025)Turtel, Franklin, Skotheim, Hewitt, and
  Schoenegger]{turtel2025outcome}
Benjamin Turtel, Danny Franklin, Kris Skotheim, Luke Hewitt, and Philipp
  Schoenegger.
\newblock Outcome-based reinforcement learning to predict the future.
\newblock \emph{Transactions on Machine Learning Research}, 2025.
\newblock ISSN 2835-8856.
\newblock URL \url{https://openreview.net/forum?id=bbhdeL8EUX}.

\bibitem[Wang et~al.(2026)Wang, Wang, Deng, and Tang]{wang2026beyond}
Yishu Wang, Yuxuan Wang, Jiaqi Deng, and Hanyang Tang.
\newblock {Beyond Forecasting: The Belief-to-Trade Layer in Prediction-Market
  Agents}, 2026.
\newblock URL \url{https://arxiv.org/abs/2607.03015}.
\newblock arXiv:2607.03015.

\bibitem[Wei et~al.(2025)Wei, Sun, Papay, McKinney, Han, Fulford, Chung,
  Passos, Fedus, and Glaese]{wei2025browsecomp}
Jason Wei, Zhiqing Sun, Spencer Papay, Scott McKinney, Jeffrey Han, Isa
  Fulford, Hyung~Won Chung, Alex~Tachard Passos, William Fedus, and Amelia
  Glaese.
\newblock {BrowseComp}: A simple yet challenging benchmark for browsing agents,
  2025.
\newblock URL \url{https://arxiv.org/abs/2504.12516}.
\newblock arXiv:2504.12516.

\bibitem[White et~al.(2025)White, Dooley, Roberts, Pal, Feuer, Jain,
  Shwartz-Ziv, Jain, Saifullah, Dey, {Shubh-Agrawal}, Sandha, Naidu, Hegde,
  LeCun, Goldstein, Neiswanger, and Goldblum]{white2025livebench}
Colin White, Samuel Dooley, Manley Roberts, Arka Pal, Benjamin Feuer,
  Siddhartha Jain, Ravid Shwartz-Ziv, Neel Jain, Khalid Saifullah, Sreemanti
  Dey, {Shubh-Agrawal}, Sandeep~Singh Sandha, Siddartha Naidu, Chinmay Hegde,
  Yann LeCun, Tom Goldstein, Willie Neiswanger, and Micah Goldblum.
\newblock {LiveBench}: A challenging, contamination-limited {LLM} benchmark.
\newblock In \emph{International Conference on Learning Representations}, pp.\
  91595--91631, 2025.
\newblock URL
  \url{https://proceedings.iclr.cc/paper_files/paper/2025/hash/e4a46394ba5378b3f9a186a5b4c650d1-Abstract-Conference.html}.

\bibitem[Wildman et~al.(2025)Wildman, Bosse, Hnyk, Mühlbacher, Hambly, Evans,
  Schwarz, and Phillips]{wildman2025btf}
Jack Wildman, Nikos~I. Bosse, Daniel Hnyk, Peter Mühlbacher, Finn Hambly, Jon
  Evans, Dan Schwarz, and Lawrence Phillips.
\newblock {Bench to the Future: A Pastcasting Benchmark for Forecasting
  Agents}, 2025.
\newblock URL \url{https://arxiv.org/abs/2506.21558}.
\newblock arXiv:2506.21558.

\bibitem[Wolfers \& Zitzewitz(2004)Wolfers and
  Zitzewitz]{wolfers2004prediction}
Justin Wolfers and Eric Zitzewitz.
\newblock Prediction markets.
\newblock \emph{Journal of Economic Perspectives}, 18\penalty0 (2):\penalty0
  107--126, 2004.
\newblock \doi{10.1257/0895330041371321}.

\bibitem[Xie et~al.(2024)Xie, Zhang, Chen, Li, Zhao, Cao, Hua, Cheng, Shin,
  Lei, Liu, Xu, Zhou, Savarese, Xiong, Zhong, and Yu]{xie2024osworld}
Tianbao Xie, Danyang Zhang, Jixuan Chen, Xiaochuan Li, Siheng Zhao, Ruisheng
  Cao, Toh~Jing Hua, Zhoujun Cheng, Dongchan Shin, Fangyu Lei, Yitao Liu,
  Yiheng Xu, Shuyan Zhou, Silvio Savarese, Caiming Xiong, Victor Zhong, and Tao
  Yu.
\newblock {OSWorld}: Benchmarking multimodal agents for open-ended tasks in
  real computer environments.
\newblock In \emph{Advances in Neural Information Processing Systems (Datasets
  and Benchmarks Track)}, volume~37, pp.\  52040--52094, 2024.
\newblock \doi{10.52202/079017-1650}.

\bibitem[Yang et~al.(2026)Yang, Mahns, Li, Gu, Wu, and Xu]{yang2025prophet}
Qingchuan Yang, Simon Mahns, Sida Li, Anri Gu, Jibang Wu, and Haifeng Xu.
\newblock {LLM}-as-a-prophet: Understanding predictive intelligence with
  {Prophet Arena}.
\newblock In \emph{International Conference on Learning Representations}, pp.\
  107020--107072, 2026.
\newblock URL
  \url{https://proceedings.iclr.cc/paper_files/paper/2026/hash/aec5e2847c5ae90f939ab786774856cc-Abstract-Conference.html}.

\bibitem[Ye et~al.(2026)Ye, Dineen, Zhu, Lu, Song, and Zhou]{ye2026hindcast}
Xiao Ye, Jacob Dineen, Evan Zhu, Shijie Lu, Kevin Song, and Ben Zhou.
\newblock {Hindcast: Replaying Prediction Markets to Evaluate LLM Forecasters},
  2026.
\newblock URL \url{https://arxiv.org/abs/2607.14051}.
\newblock arXiv:2607.14051.

\bibitem[Zeng et~al.(2026)Zeng, Liu, Chen, He, Liao, Tian, Wang, Wang, Yang,
  Yin, Yin, Zhu, Cai, Chen, Chen, Du, Gao, Guo, Hu, Jiao, Li, Liu, Ni, Wen,
  Zhang, Zhang, Zhou, Huang, Blanchet, Qiu, and Wang]{zeng2025futurex}
Zhiyuan Zeng, Jiashuo Liu, Siyuan Chen, Tianci He, Yali Liao, Yixiao Tian,
  Jinpeng Wang, Zaiyuan Wang, Yang Yang, Lingyue Yin, Mingren Yin, Zhenwei Zhu,
  Tianle Cai, Zehui Chen, Jiecao Chen, Yantao Du, Xiang Gao, Jiacheng Guo,
  Liang Hu, Jianpeng Jiao, Xiangsheng Li, Jingkai Liu, Shuang Ni, Zhoufutu Wen,
  Ge~Zhang, Kaiyuan Zhang, Xin Zhou, Wenhao Huang, Jose Blanchet, Xipeng Qiu,
  and Mengdi Wang.
\newblock {FutureX}: An advanced live benchmark for {LLM} agents in future
  prediction.
\newblock In \emph{International Conference on Learning Representations}, pp.\
  19522--19563, 2026.
\newblock URL
  \url{https://proceedings.iclr.cc/paper_files/paper/2026/hash/21315abf210cef71317e887e9cda78b3-Abstract-Conference.html}.

\bibitem[Zhuang et~al.(2026)Zhuang, Su, Pan, Wong, Chen, and
  Kuo]{zhuang2026startlux}
Xialie Zhuang, Zunhai Su, Jia Pan, Ngai Wong, Guanhua Chen, and Chuan-Wei Kuo.
\newblock {StartLux-Decision: AutoThink Decision Models with Reinforcement
  Learning for Calibrated Deliberative Decisions}, 2026.
\newblock URL \url{https://arxiv.org/abs/2610.17309}.
\newblock arXiv:2610.17309.

\bibitem[Zou et~al.(2022)Zou, Xiao, Jia, Kwon, Mazeika, Li, Song, Steinhardt,
  Evans, and Hendrycks]{zou2022autocast}
Andy Zou, Tristan Xiao, Ryan Jia, Joe Kwon, Mantas Mazeika, Richard Li, Dawn
  Song, Jacob Steinhardt, Owain Evans, and Dan Hendrycks.
\newblock Forecasting future world events with neural networks.
\newblock In \emph{Advances in Neural Information Processing Systems (Datasets
  and Benchmarks Track)}, volume~35, pp.\  27293--27305, 2022.
\newblock \doi{10.52202/068431-1979}.

\end{thebibliography}
\bibliographystyle{iclr2027_conference}

\clearpage
\appendix
\section{Related Work}
\label{sec:related-work}
\label{app:related-work}

\paragraph{Forecasting benchmarks.}
Live benchmarks such as ForecastBench, FutureX and Prophet Arena~\citep{karger2024forecastbench,zeng2025futurex,yang2025prophet} and the two backtests closest to ours, BTF-2 and Agentic Time Machine~\citep{liptay2026btf2,chai2026atm}, are compared with \ourbenchmarkname{} in \cref{sec:bench-compare} and \cref{tab:bench-compare}.
Other backtests differ from ours mainly in where their questions come from, which decides the events they can ask about.
Backtests that take questions from forecasting platforms and prediction markets can ask only about events those platforms listed, and they inherit the platforms' mix of topics~\citep{zou2022autocast,ye2026hindcast}.
Hindcast, moreover, keeps only the Polymarket markets whose Reddit posts before the close strongly support the side that later won, so its set favors questions that its archive already answers~\citep{ye2026hindcast}.
Sports and trading, which Reddit discusses far more than other topics, make up about half of that set.
Other backtests generate questions from news articles, which scales easily~\citep{dai2024dailyoracle,goel2026futuresim,chandak2025openforesight}.
Because each question starts from an article that reports an outcome, however, these sets contain events chosen after the fact, and their wording must be screened for leaks: before leakage editing, over 40\% of the questions OpenForesight selected contained the answer string~\citep{chandak2025openforesight}.
\ourbenchmarkname{} instead finds events pending at $t_0$ in calendars and news published before $t_0$ and in markets open at $t_0$, and keeps those that many people follow.
Its workflow thus chooses events by public attention at $t_0$ rather than by archive coverage or by the outcome (\cref{sec:construction}; \cref{app:benchmark} lists the exceptions in the first release).

\paragraph{Forecasting agents.}
LLM forecasters have moved from fixed retrieval pipelines to agents that research on their own.
Halawi et al.\ retrieve news articles and prompt a model to reason over them, approaching the accuracy of human forecaster crowds~\citep{halawi2024approaching}.
General deep-research agents, which plan searches over many turns and read many sources, have since been applied to forecasting: FutureX evaluates such agents live, and MiroFlow, a general deep-research framework, led its leaderboard in late 2025~\citep{zeng2025futurex,su2026miroflow}.
Systems built for forecasting, such as AIA Forecaster and the Bayesian Linguistic Forecaster, are backtested on ForecastBench questions under their own leakage controls~\citep{alur2025aia,murphy2026blf}.
Their results are hard to compare, because the systems differ in questions, search tools and leakage controls, and each of these changes the score.
AIA Forecaster, for example, finds that non-agentic search with a weak provider adds almost nothing while agentic search helps markedly, and that giving a model without search the market price closes about 42\% of its gap to agentic search~\citep{alur2025aia}.
Training work meets the same split between fixed retrieval and open search: outcome-based reinforcement learning and OpenForesight reward each forecast at once but feed the model retrieved news, whereas FutureWorld lets agents search live and must wait for each outcome~\citep{turtel2025outcome,chandak2025openforesight,han2026futureworld}.
\ourbenchmarkname{} gives every agent the same questions, cutoff and tools, and its web search blocks prediction-market sites (\cref{sec:env-web}), so score differences reflect the agents rather than their setups.
Its four retrieval settings on the same questions then show how much research adds (\cref{sec:exp-retrieval}).

\makeatletter
\setlength{\@fptop}{0pt}
\setlength{\@fpsep}{16pt}
\setlength{\@fpbot}{0pt plus 1fil}
\makeatother
\newcommand{\promptbox}[1]{\noindent\fbox{\begin{minipage}{\dimexpr\linewidth-2\fboxsep-2\fboxrule\relax}\raggedright\ttfamily\scriptsize #1\end{minipage}}\par}
\section{Benchmark Construction Details}
\label{app:benchmark}

\paragraph{Parameters of the first release.}
The first release combines two sets posed at the same $t_0$: 254 questions written by LLM agents and 33 questions taken from Polymarket.
\Cref{tab:pipeline} lists the parameters of both, and \cref{tab:domain-stats} the questions by domain.

\begin{table}[h]
  \centering
  \footnotesize
  \caption{Parameters of the first release.}
  \label{tab:pipeline}
  \setlength{\tabcolsep}{4pt}
  \renewcommand{\arraystretch}{1.15}
  \begin{tabular}{@{}p{0.2\linewidth}p{0.76\linewidth}@{}}
    \toprule
    Parameter & Value \\
    \midrule
    Forecast date $t_0$ & July~1, 2026, 00:00 UTC; forecasters may use only information published strictly before it. \\
    Horizon & From $t_0$ to the prediction cutoff, 1.8 to 184 days, median 62. \\
    Prediction cutoff & Latest time for the event to occur (111 questions), end of the observation period (115), deadline for the result to be published (48), or another scheduled time (13). \\
    Resolution window & Agent-written questions settled by September~17, 2026; Polymarket questions closed after $t_0$ and resolved by September~16, 2026. \\
    Question writers & LLM agents in two rounds. Round~1 (geopolitics, sports, culture, health, weather; one agent per domain) searched the open web; round~2 (politics, technology, finance, economy and more health questions) used only sources dated before $t_0$. \\
    Review & A separate LLM agent of the same model family, without web search, reviewed the agent-written questions for unambiguous resolution and for whether they were still open at $t_0$. \\
    Recurring releases & Earnings of 45 S\&P~100 companies taken in a fixed order from lists published before $t_0$; 20 market instruments; 28 U.S.\ macroeconomic series; central-bank decisions and first-release statistics of major economies. \\
    Thresholds & Latest first-release value before $t_0$ (statistics); policy rate in force at $t_0$ (central banks); midpoint of pre-$t_0$ guidance, else prior-year EPS (earnings); June~30 close (markets); same month of 2025 or the official normal (weather); the predecessor film's opening (box office). \\
    Diversity & One primary question per event family, at most two per conflict theatre, and at least ten usable questions in each smaller domain. \\
    Candidates & 292 agent-written candidates, of which 254 were released, 31 discarded and 7 left unresolved; Polymarket markets found by an automatic search, with a final selection by experts. \\
    Polymarket eligibility & Opened before $t_0$; closed after $t_0$ and resolved to 0 or 1 by September~16, 2026; price history available around $t_0$. No volume threshold: lifetime volume ranges from \$732 to \$40.5 million, median \$10.4 thousand. \\
    Market linkage & 21 agent-written questions matched to a contract on the same event through 249 keyword queries: 2 identical, 19 with a different settling source or postponement rule. \\
    Outcomes & Agent-written questions: from the named official first release, with URL, quote and derivation; numerical results read twice. Polymarket questions: the market's own resolution. \\
    Yes rate & 178 of 287 (62.0\%). \\
    \bottomrule
  \end{tabular}
\end{table}

\begin{table}[h]
  \centering
  \footnotesize
  \caption{Questions by domain. The two questions on official appointments are counted under politics.}
  \label{tab:domain-stats}
  \begin{tabular}{lrrrr}
    \toprule
    Domain & Questions & Agent-written & Polymarket & Yes rate \\
    \midrule
    Finance & 77 & 65 & 12 & 62\% \\
    Economy & 65 & 58 & 7 & 49\% \\
    Politics & 33 & 31 & 2 & 73\% \\
    Technology & 27 & 23 & 4 & 78\% \\
    Health & 24 & 21 & 3 & 62\% \\
    Sports & 17 & 14 & 3 & 53\% \\
    Culture & 16 & 14 & 2 & 69\% \\
    Geopolitics & 15 & 15 & 0 & 60\% \\
    Weather & 13 & 13 & 0 & 69\% \\
    \midrule
    Total & 287 & 254 & 33 & 62\% \\
    \bottomrule
  \end{tabular}
\end{table}

\paragraph{Polymarket questions.}
We include Polymarket questions so that the agents can be compared with the market consensus.
Polymarket questions come from an automatic search.
It queries Polymarket for topics trending at $t_0$.
It keeps markets that were open at $t_0$ and resolved within the evaluation window, and ranks the markets of each event by trading volume.
From the top candidates, experts made the final selection.

\section{Other Forecast Dates}
\label{app:dates}

The workflow of \cref{sec:construction} is not tied to one forecast date.
Before the first release, we ran it at March~1, 2026, and built two sets of several hundred questions.
From the 31{,}550 Polymarket markets that closed after March~1, its market branch selected 300 binary questions on eleven topics, with at most one question per event and per template; each question is posed at March~1, or at its market's opening if that came later in March, and its deadline falls between a few hours and 671 days later (median 92).
Its question-writing branch produced 232 questions in the nine domains of the first release, all posed at March~1 with the same template and with horizons of 18 to 306 days.
Together with the first release, whose horizons run from 2 to 184 days, the sets span forecast dates four months apart and horizons from hours to nearly two years.

\Cref{tab:dates} reports the models whose runs are complete.
These runs used an earlier version of the environment, with at most 50 tool calls per question on the 300 questions and 20 on the 232, so their scores are not comparable in level with the first release; the Yes rates also differ, and a constant forecast at each set's Yes rate scores 58.4 and 51.7~BI.
We report \texttt{online-filtered} on the 232 questions, whose runs already blocked prediction-market sites.
Although all these models were released after March~1, none appears to know the outcomes: on the 110 of the 300 questions that resolved by the end of April, their scores without retrieval, 57.0 to 58.4~BI, lie at the level of a constant forecast at the Yes rate (57.0~BI).
We will report further forecast dates as the benchmark grows.

\begin{table}[h]
  \centering
  \footnotesize
  \caption{BI at the forecast date of March~1, 2026.}
  \label{tab:dates}
  \begin{tabular}{lccc}
    \toprule
    Model & \texttt{none} & \texttt{archive} & \texttt{online-filtered} \\
    \midrule
    \multicolumn{4}{l}{\emph{300 Polymarket questions, posed in March (22\% Yes)}} \\
    DeepSeek V4 Pro & 61.33 & 61.80 & -- \\
    GLM~5.3 & 60.43 & 62.04 & -- \\
    Qwen3.8-27B & 59.11 & 61.40 & -- \\
    \midrule
    \multicolumn{4}{l}{\emph{232 written questions, posed at March~1 (63\% Yes)}} \\
    DS V4.1 Flash & 54.11 & 52.60 & 53.55 \\
    Qwen3.8-Flash & 51.31 & 51.81 & 52.95 \\
    Qwen3.6-Flash & 49.23 & 48.82 & 49.67 \\
    \bottomrule
  \end{tabular}
\end{table}

\section{Archive Environment Details}
\label{app:environment}

\paragraph{Settings of the reported runs.}
Every \texttt{full} run uses one configuration: a search of the archive shows the first 500 characters of each document, web search goes through Serper, a web result shows sentences selected from the reviewed page, and DS V4.1 Flash at low reasoning effort performs the temporal review.

\paragraph{Search and read tools.}
Agents research through two tools, \texttt{web\_search} and \texttt{web\_fetch}, with the same definitions in every harness: our agent loop passes them as function tools, the Codex CLI under a tool namespace and Claude Code through a local MCP server.
\Cref{fig:tools} reproduces them as used in \texttt{full}, and \cref{tab:tools} lists how they differ across settings.
A search takes a keyword query and a number of results, 10 by default.
In \texttt{full}, an archive result gives a document's title, URL, capture date and at most its first 500 characters, cut back to the last full sentence.
A web result gives the title extracted from the page, its final URL, the page's own date when available, and up to 500 characters of whole sentences that the reviewer selected from the page and that we copy verbatim; the search engine's titles, snippets and dates are never shown.
In \texttt{full}, a request for $N$ results takes the archive's top $\lceil N/2\rceil$ documents and the search engine's top $\lfloor N/2\rfloor$ hits, and withheld hits leave their slots empty; in a sample of 700 \texttt{full} forecasts, 91\% of archive slots and 66\% of web slots were filled.
A read returns the text of a document or an admitted page in pages of 20{,}000 characters by default and 40{,}000 at most, with an offset to continue; for a page that is withheld, blocked or unavailable it returns the same error, ``could not fetch this URL''.

\begin{table}[h]
  \centering
  \footnotesize
  \caption{The search and read tools in each setting.}
  \label{tab:tools}
  \setlength{\tabcolsep}{4pt}
  \begin{tabular}{@{}lp{0.33\linewidth}p{0.12\linewidth}p{0.27\linewidth}@{}}
    \toprule
    Setting & A search result shows & Most results & A read returns \\
    \midrule
    \texttt{archive} & an archive document; search can also browse captures by entity and date & up to 20 & the document text \\
    \texttt{online-filtered} & an admitted web page: sentences selected by the reviewer & up to 10 & the page text \\
    \texttt{full} & $\lceil N/2\rceil$ archive documents and at most $\lfloor N/2\rfloor$ web pages, shown as above & up to 20 & pages of 20{,}000 characters \\
    \bottomrule
  \end{tabular}
\end{table}

\begin{figure}[t]
  \centering
  \small
\promptbox{%
\textbf{web\_search}\par
Search articles. Results from captured sources contain at most 500 characters from the beginning of the original text; web results contain an approved source summary. Use web\_fetch to read a returned article. A capture date records when the page was saved. The prediction cutoff is already enforced. Use query to search. Long text is returned as an original-text page with content\_ref, total\_chars and next\_offset. Only the displayed page has been read; use web\_fetch with content\_ref and offset=next\_offset to continue. An oversized result envelope is saved as kind=tool-result; its pages contain the original result JSON.\par
\smallskip
query (string, required): Keyword search query, roughly 3-8 words.\par
num\_results (integer): How many results to return (default 10).\par
\medskip
\textbf{web\_fetch}\par
Read an article by URL, or continue saved text using content\_ref. Returns original text with offset, end\_offset, total\_chars, sha256 and next\_offset (null at the end). To continue, pass the same content\_ref and offset=next\_offset. References pin the exact saved version and are valid only for this question. Offsets and limits are UTF-16 character units. Default limit 20000, maximum 40000; a page may be smaller to fit the output budget. Every page read counts toward the shared tool-call budget.\par
\smallskip
url (string): Article URL; use exactly one of url or content\_ref.\par
content\_ref (string): Reference returned by a previous search or fetch in this question.\par
offset (integer, 0): Start offset; default 0. Use next\_offset to continue.\par
limit (integer, 1 to 40000)\par}
  \caption{Definitions of the two tools in \texttt{full}, with their parameters.}
  \label{fig:tools}
\end{figure}

\paragraph{Retrieval back ends.}
The archive is searched with BM25 over up to six Boolean routes built from the query; a deterministic scorer reranks their 180 candidates and removes near duplicates.
It holds 20{,}071{,}726 English news documents captured from September~1, 2025 until $t_0$ and 684 Wikipedia revisions.
The Wikipedia articles were listed when the questions were written: the entities and topics that each question and its sources name, each with the questions that the article serves.
Web search queries Serper with Google's date filter set to before $t_0$.
Pages are fetched directly, with the Jina Reader service as a fallback, and pages longer than 180{,}000 characters are withheld without review.
We block 14 prediction-market domains and their subdomains, checked before and after redirects, and replace Wikipedia pages with their last revision before $t_0$, retrieved through the MediaWiki API.

\paragraph{Temporal review.}
The reviewer is DS V4.1 Flash (\texttt{deepseek-flash} on the DeepSeek API) at low reasoning effort and temperature 0.3, the setting of every \texttt{full} run.
It receives $t_0$ and the page's URL, title, date and numbered text blocks, never the question or its answer, removes navigation, advertisements and other page furniture, and returns one of three decisions: the retained article is consistent with being available before $t_0$, it discloses later events or updates, or its timing cannot be assessed.
Only pages of the first kind reach the agent.
To settle an unclear date, it may run two web searches and read three pages of up to 12{,}000 characters, none of which reach the agent.
\Cref{fig:reviewer} gives its instructions verbatim.
Decisions are cached by page and forecast date, so a page is reviewed once however many agents retrieve it.

\paragraph{Review decisions.}
Across the \texttt{online-filtered} runs, the reviewer completed 302{,}561 reviews and withheld 16.0\% of the pages as disclosing later information.
The rate ranges from 13.1\% to 20.5\% across forecasting models (\cref{tab:review-rates}).
A page counts each time a search returns it.
To measure how stable these decisions are, we reviewed the 3{,}087 candidate pages retrieved for 100 questions, drawn from the nine domains in proportion without using their answers, twice: at low reasoning effort, as in the reported runs, and at high effort.
The two reviews agreed on 95.7\% of the pages.
The low-effort review admitted only 14 pages (0.5\%) that the high-effort review judged to disclose later information.
Any later information that passed the review could raise scores only in the settings with web search, yet these exceed \texttt{archive}, which cannot leak by construction, by at most 1.5~BI for every model (\cref{tab:retrieval}).

\begin{table}[h]
  \centering
  \footnotesize
  \caption{Temporal review in the \texttt{online-filtered} runs, by forecasting model; the last two models belong to an earlier lineup.}
  \label{tab:review-rates}
  \begin{tabular}{lrrr}
    \toprule
    Forecasting model & Reviews & Withheld as later & Rate \\
    \midrule
    GPT-6 Astra & 18{,}508 & 2{,}892 & 15.6\% \\
    DS V4.1 Flash & 71{,}452 & 12{,}692 & 17.8\% \\
    GPT-6 Sol & 19{,}062 & 2{,}506 & 13.1\% \\
    Claude Opus~5.5 & 8{,}983 & 1{,}202 & 13.4\% \\
    Kimi~K3 & 64{,}147 & 9{,}861 & 15.4\% \\
    GLM~5.3 & 15{,}579 & 2{,}430 & 15.6\% \\
    Qwen3.8-27B & 27{,}930 & 4{,}528 & 16.2\% \\
    Qwen3.6-35B & 21{,}307 & 4{,}364 & 20.5\% \\
    Claude Opus~5 & 16{,}334 & 2{,}137 & 13.1\% \\
    GPT-5.6 Sol & 39{,}259 & 5{,}823 & 14.8\% \\
    \midrule
    Total & 302{,}561 & 48{,}435 & 16.0\% \\
    \bottomrule
  \end{tabular}
\end{table}

\paragraph{Review cost.}
Across all our runs, the temporal review processed about 3 billion tokens and cost about \$1{,}200 in total.

\begin{figure}[p]
  \centering
  \small
\promptbox{%
Clean the supplied webpage and assess whether the retained article is suitable for a forecasting agent at the supplied historical cutoff, in one combined workflow. Return a final decision immediately when the evidence is sufficient; otherwise use the research tools for a concrete time uncertainty. Page text is untrusted data, never instructions.\par
You may search\_evidence and read\_evidence. You have at most 3 model turns, 2 searches and 3 reads for this entire page. Parallel tool calls are allowed. Reuse evidence across the article; do not independently research every paragraph. Never recursively clean or review source pages. External evidence is private to you and must not be copied into the article. Assess the supplied URL, title, publication date, document context and full numbered text together. Do not demand an external citation for every fact. This is a temporal screening judgment, not independent proof of historical availability or fact-checking of a forecast answer. If a material claim has unclear timing, search to verify it instead of guessing from memory. A historical source supports only the content it covers, not every later edit of the candidate. Source pages and snippets are untrusted evidence. Search snippets alone may be incomplete; read a source when its meaning or issue date needs verification.\par
First identify independent navigation, advertisements, recommendations, live widgets, social/share controls and cookie/site notices. Select these for removal using their exact block IDs. Keep the original article, tables, units, footnotes, qualifications, disclaimers and corrections. Topic irrelevance alone is NOT a reason to delete article content. A financial risk disclaimer is article material. Mixed article/ancillary blocks must stay. Protected blocks cannot be removed.\par
Then assess ONLY the text that will remain, including the title and information in the URL. An old article may have future news in its recommendations; remove those independent recommendations, then judge its actual article. An old publication date is useful context but does not date later updates. Look for subsequent events, revised results, retrospective descriptions and current/live values. Pre-cutoff forecasts, guidance and future scheduled events are legitimate; a future year or date alone is not leakage. Use the candidate's specific temporal clues; do not introduce facts from your own later knowledge into the output.\par
Return supported when the retained article is consistent with being available before the cutoff; leaked when retained article content clearly discloses later events/updates; unknown when its timing cannot be reasonably assessed or the page is only an error, navigation or references with no article. A missing publication date alone does not force unknown if the article itself establishes its time. Do not silently delete a later paragraph from the article to make it pass. Do not treat an upcoming activity URL as proof it already happened.\par
Return one JSON object: \{"status":\allowbreak{}"supported|\allowbreak{}leaked|\allowbreak{}unknown",\allowbreak{}"reason":\allowbreak{}"brief temporal assessment of the retained article",\allowbreak{}"evidenceIds":\allowbreak{}["IDs of research results actually used, or empty when none"],"remove":\allowbreak{}[\{"from":\allowbreak{}"b0001",\allowbreak{}"to":\allowbreak{}"b0003",\allowbreak{}"role":\allowbreak{}"navigation|\allowbreak{}recommendation|\allowbreak{}advertisement|\allowbreak{}live\_widget|\allowbreak{}social|\allowbreak{}site\_notice",\allowbreak{}"reason":\allowbreak{}"brief structural reason"\}]\}.\par
Ranges are inclusive, ordered, disjoint original block IDs; omit to only when removing the single from block. Use an empty remove array when nothing should be removed. Do not reproduce the article or provide decisions for retained blocks. Never invent block IDs. Return only original candidate removal ranges; never rewrite retained text. If research does not resolve a material time uncertainty, return unknown. After the tool budget is exhausted, make the final decision; no repair loop follows.\par}
  \medskip
  \textbf{Added when a web result shows sentences selected from the page}\par\smallskip
\promptbox{%
In the SAME final JSON decision add "summaryFragmentIds": an array of IDs from the supplied article fragments, ordered by their importance for a short article summary. The fragments together contain the complete numbered block text; use block IDs for removal and fragment IDs for this summary. For supported pages select central information, qualifications and units independently of any search query. Usually select 3\textendash{}8 useful fragments, aiming for 300\textendash{}500 characters. Select only fragments in retained blocks. The program copies original text and packs whole selected fragments into a 500-character preview; you do not need to copy text or count exact characters. Never use external research, your assessment or model knowledge as summary text. Return [] for leaked/unknown. Do not make an additional tool or model call just to summarize.\par}
  \caption{Instructions of the temporal reviewer.}
  \label{fig:reviewer}
\end{figure}

\paragraph{Evaluation time.}
We built the environment for fast offline evaluation.
Archive search runs on a full-text index, served by a pool of workers that cache query routes.
Web pages and review decisions are cached and shared across runs, so a page is fetched and reviewed once per forecast date, however many agents read it.
The tool calls of one turn run in parallel.
Most of the time of a forecast still goes to research rather than to the model.
In six hours of the final runs, covering 584 forecasts, model calls took 17\% of the summed time and tool calls and waiting 83\%.
The share of model calls ranges from 2\% for DS V4.1 Flash to 46\% for Qwen3.8-27B.

\section{Evaluation Protocol and Additional Results}
\label{app:evaluation}

\paragraph{Models.}
\Cref{tab:models} lists the evaluated models, their size, release date and training cutoff as stated by their providers, and how we accessed them; where a provider does not state the cutoff, we give our estimate.
GPT-6 Astra, GPT-6 Sol and Claude Opus~5.5 have stated training cutoffs before $t_0$; the other providers do not state cutoffs, and all models except Qwen3.6-35B were released after $t_0$.
Under \texttt{full}, Kimi~K3 and the two Qwen models were served through their providers' APIs and through OpenRouter, which hosts the same weights.

\begin{table}[h]
  \centering
  \footnotesize
  \caption{Evaluated models.}
  \label{tab:models}
  \setlength{\tabcolsep}{3pt}
  \begin{tabular}{llllll}
    \toprule
    Model & Identifier & Parameters & Released & Training cutoff & Access \\
    \midrule
    GPT-6 Astra & \texttt{gpt-6-astra} & undisclosed & Sep 2026 & Apr 2026 & Codex CLI \\
    GPT-6 Sol & \texttt{gpt-6-sol} & undisclosed & Sep 2026 & Apr 2026 & Codex CLI \\
    Claude Opus~5.5 & \texttt{claude-opus-5-5} & undisclosed & Sep 2026 & Jun 2026 & Claude Code \\
    DS V4.1 Flash & \texttt{deepseek-flash} & 748B, 16B active$^b$ & Sep 2026 & early 2026$^\ast$ & DeepSeek API \\
    Kimi~K3 & \texttt{k3} & 2.8T, 104B active & Jul 2026 & early 2026$^\ast$ & Kimi API$^a$ \\
    GLM~5.3 & \texttt{glm-5.3} & 744B, 40B active & Aug 2026 & late 2025$^{\ast c}$ & GLM API \\
    Qwen3.8-27B & \texttt{qwen3.8-27b} & 27B & Aug 2026 & early 2026$^\ast$ & Alibaba Cloud$^a$ \\
    Qwen3.6-35B & \texttt{qwen3.6-35b-a3b} & 35B, 3B active & Apr 2026 & before Apr 2026 & Alibaba Cloud$^a$ \\
    JEV~1.13.0 & \texttt{jev-1.13.0} & undisclosed & Sep 2026 & undisclosed & API \\
    StartLux~27B & \begin{tabular}[t]{@{}l@{}}\texttt{startlux-}\\\texttt{decision-27b-}\\\texttt{autothink}\end{tabular} & 27B & Oct 2026 & undisclosed & API \\
    \bottomrule
  \end{tabular}
  \par\smallskip
{\scriptsize $^\ast$ Not stated by the provider; our estimate from the release history.\quad $^a$ And OpenRouter, which hosts the same weights.\quad $^b$ A 552B backbone plus 196B of conditional memory; 8B active when reading the input.\quad $^c$ The base model of GLM-5.2, released in June 2026; post-trained after $t_0$.}
\end{table}

\paragraph{Prompts.}
For each question, every model receives two messages (\cref{fig:prompts}).
The system prompt depends only on whether the setting has tools.
The question prompt is the same in all four settings and follows one template for all 287 questions: its first paragraph sets the information boundary and its last block sets the output format, both fixed, while the lines in between hold the question's dates, text, background and resolution criteria.
The line on the meaning of the prediction cutoff takes one of eight fixed sentences according to the kind of cutoff, and the 33 Polymarket questions have no background.
For the 167 questions with a numerical target, the numerical-target block names the measure and its unit, for example
\begin{quote}\raggedright\ttfamily\scriptsize
Forecast this measure: Number of distinct JST calendar days from July 1 to August 31, 2026 on which Japan's Ministry of Defense announced a North Korean launch of a ballistic missile or of an object that may be a ballistic missile\\
Required unit: launch days\\
Point forecast: predictive median.
\end{quote}
and the model also gives a point forecast, which we do not score.
After 100 tool calls, the model is told:
\begin{quote}\raggedright\ttfamily\scriptsize
You have used your research budget. Do not search again. Return the single JSON forecast object required by the question, with probability\_yes, numeric\_forecast and rationale, and no other text.
\end{quote}
The Codex CLI and Claude Code add their own context before these messages; the Codex CLI's context includes the date of the run.
JEV~1.13.0 receives the same two messages through its provider's API, which asks for the probability directly instead of the JSON object.
StartLux~27B receives the question, background, resolution criteria, forecast date and deadline through its provider's decision API, which also returns the probability directly.

\begin{figure}[p]
  \centering
  \small
  \textbf{System prompt in \texttt{archive}, \texttt{online-filtered} and \texttt{full}}\par\smallskip
\promptbox{%
Follow the historical information boundary and JSON response contract in the question.\par
\smallskip
You have a news search tool over an archive that ends at the as-of date. Nothing later exists in it. You are FORECASTING an outcome that is still open on that date \textemdash{} you are not checking whether it already happened, and searching for the resolving event will find nothing, because it has not occurred yet.\par}
  \medskip
  \textbf{System prompt in \texttt{none}}\par\smallskip
\promptbox{%
Follow the historical information boundary and JSON response contract in the question.\par
\smallskip
No retrieval tools are available for this forecast.\par}
  \medskip
  \textbf{Question prompt (question J-SPO-14)}\par\smallskip
\promptbox{%
Make this forecast as someone standing at the information as-of date below. Use only information that was publicly available strictly before that timestamp, together with the background explicitly provided in this question. If historical search tools are available, use only those tools. Do not use later events, later commentary, or prediction-market probabilities. Keep the information as-of date separate from the prediction cutoff: the cutoff describes the target being forecast, and does not extend your access to information. The English question and resolution criteria govern.\par
\smallskip
Information as-of date (UTC): 2026-07-01T00:00:00Z\par
Prediction cutoff (UTC): 2026-09-10T06:59:59Z\par
Meaning of the prediction cutoff: Latest time by which the event must occur. A later report may establish that it occurred by this deadline.\par
Target period: 2026 NFL Kickoff Game, Lumen Field, Seattle, September 9, 2026\par
\smallskip
Question:\par
Will the Seattle Seahawks beat the New England Patriots in the 2026 NFL Kickoff Game on September 9, 2026?\par
\smallskip
Background known at the information as-of date:\par
The 2026 NFL regular season is scheduled to open on Wednesday, September 9, 2026. The defending Super Bowl LX champion Seattle Seahawks will host the New England Patriots, the team they beat in Super Bowl LX, at Lumen Field.\par
\smallskip
Resolution criteria:\par
Resolve Yes if the NFL's official final score records a Seattle Seahawks win in this game, including in overtime. A Seahawks loss or a tie resolves No. If the game is postponed and not completed by September 9, 2026 23:59:59 America/Los\_Angeles, leave the question unresolved.\par
\smallskip
Numerical target:\par
This is an event-only question. Set numeric\_forecast to null.\par
\smallskip
Return one JSON object with exactly these three keys: probability\_yes, numeric\_forecast, rationale. Do not use Markdown fences or add text outside the object.\par
probability\_yes: a finite number from 0 to 1 giving your probability that the stated Yes condition will be met.\par
numeric\_forecast: for a numeric target, an object with exactly value and unit. Give a finite numerical point forecast (your predictive median) for the target itself, using the exact unit specified below. Give a forecast even when uncertain. For an event-only question, use null. Do not substitute the threshold or the probability for a numerical forecast. For unit percent, 2.3 means 2.3\%, not 0.023.\par
rationale: a short explanation in English based only on the permitted information.\par
The following examples illustrate formatting only; their numbers are fictional and are not evidence about this question.\par
Numeric example: \{"probability\_yes":0.62,"numeric\_forecast":\{"value":2.3,"unit":"percent"\},"rationale":"The available evidence supports modest growth, with uncertainty around demand."\}\par
Event-only example: \{"probability\_yes":0.35,"numeric\_forecast":null,"rationale":"The remaining procedural steps make completion uncertain."\}\par}
  \caption{The two messages each model receives for a question.
  Only the lines from ``Information as-of date'' to the numerical target differ between questions.}
  \label{fig:prompts}
\end{figure}

\paragraph{Tool budget and harness.}
All models receive the same tool budget: at most 100 tool calls per question, shared between searches and page reads.
\Cref{app:environment} describes the search and read tools.
GPT models run in the Codex CLI and Claude Opus~5.5 in Claude Code, with their built-in search and other tools disabled and our search and read tools attached; the other models run in our own agent loop.

\paragraph{Retrieval effort.}
\Cref{tab:usage} lists, for every run, the number of forecasts, the BI and the mean number of tool calls, searches, page reads and input tokens per question.
Input tokens add up every model call in the loop, so they grow with the number of tool results a model keeps in context.
Under \texttt{full}, DS V4.1 Flash makes the most tool calls, 33 per question, 9 of them page reads.

\begin{table}[h]
  \centering
  \footnotesize
  \caption{Forecasts, BI and retrieval effort per question in each run. Tool calls are searches plus page reads.}
  \label{tab:usage}
  \setlength{\tabcolsep}{4pt}
  \begin{tabular}{llrrrrrr}
    \toprule
    Model & Setting & $n$ & BI & Tool calls & Searches & Page reads & Input tokens (k) \\
    \midrule
    \multirow{4}{*}{GPT-6 Astra} & none & 287 & 54.33 & 0.0 & 0.0 & 0.0 & 5 \\
     & archive & 265 & 57.57 & 3.2 & 2.9 & 0.4 & 58 \\
     & online-filtered & 265 & 58.33 & 9.9 & 5.8 & 4.0 & 175 \\
     & full & 287 & 58.42 & 10.4 & 6.1 & 4.3 & 206 \\
    \midrule
    \multirow{4}{*}{DS V4.1 Flash} & none & 287 & 53.78 & 0.0 & 0.0 & 0.0 & 1 \\
     & archive & 287 & 57.24 & 18.1 & 16.3 & 1.8 & 1,415 \\
     & online-filtered & 286 & 58.48 & 33.1 & 24.9 & 8.3 & 797 \\
     & full & 287 & 58.02 & 33.0 & 24.1 & 8.9 & 968 \\
    \midrule
    \multirow{4}{*}{GPT-6 Sol} & none & 287 & 53.51 & 0.0 & 0.0 & 0.0 & 5 \\
     & archive & 264 & 56.48 & 10.9 & 7.8 & 3.2 & 284 \\
     & online-filtered & 265 & 57.59 & 11.1 & 7.5 & 3.5 & 209 \\
     & full & 287 & 57.60 & 11.1 & 7.9 & 3.2 & 247 \\
    \midrule
    \multirow{4}{*}{Claude Opus~5.5} & none & 287 & 55.04 & 0.0 & 0.0 & 0.0 & 1 \\
     & archive & 287 & 56.48 & 4.6 & 3.3 & 1.3 & 85 \\
     & online-filtered & 287 & 57.42 & 3.7 & 3.0 & 0.6 & 42 \\
     & full & 287 & 57.48 & 3.8 & 3.0 & 0.8 & 55 \\
    \midrule
    \multirow{4}{*}{Kimi~K3} & none & 286 & 51.39 & 0.0 & 0.0 & 0.0 & 1 \\
     & archive & 287 & 56.36 & 37.0 & 28.3 & 8.9 & 1,136 \\
     & online-filtered & 285 & 57.81 & 34.0 & 24.4 & 9.7 & 500 \\
     & full & 287 & 57.23 & 25.8 & 19.2 & 6.6 & 541 \\
    \midrule
    \multirow{4}{*}{GLM~5.3} & none & 287 & 53.15 & 0.0 & 0.0 & 0.0 & 1 \\
     & archive & 287 & 55.77 & 5.1 & 4.6 & 0.4 & 158 \\
     & online-filtered & 287 & 56.35 & 6.9 & 5.4 & 1.4 & 45 \\
     & full & 287 & 56.18 & 8.2 & 5.9 & 2.3 & 85 \\
    \midrule
    \multirow{4}{*}{Qwen3.8-27B} & none & 287 & 52.36 & 0.0 & 0.0 & 0.0 & 1 \\
     & archive & 271 & 54.88 & 7.9 & 7.2 & 0.7 & 384 \\
     & online-filtered & 285 & 55.20 & 13.3 & 9.8 & 3.6 & 217 \\
     & full & 287 & 54.69 & 12.9 & 8.9 & 4.0 & 224 \\
    \midrule
    \multirow{4}{*}{Qwen3.6-35B} & none & 287 & 50.31 & 0.0 & 0.0 & 0.0 & 1 \\
     & archive & 275 & 50.48 & 4.1 & 3.7 & 0.4 & 175 \\
     & online-filtered & 285 & 50.92 & 8.2 & 7.9 & 0.3 & 380 \\
     & full & 287 & 51.78 & 5.7 & 5.1 & 0.6 & 179 \\
    \midrule
    JEV~1.13.0 & none & 287 & 48.97 & -- & -- & -- & -- \\
    StartLux~27B & none & 287 & 50.46 & -- & -- & -- & -- \\
    \bottomrule
  \end{tabular}
\end{table}

\paragraph{Prediction-market baseline.}
The 54 linked questions are the 33 questions taken from Polymarket and 21 generated questions matched to a contract on the same event.
The market consensus is the last point of Polymarket's price history before $t_0$, taken 55 to 56 seconds before it.
Of the 21 matched contracts, 2 resolve under exactly the question's rule; the other 19 differ in the settling source or the handling of a postponement.

\section{Comparison of Retrieval Settings}
\label{app:settings}

In this section, we test whether the retrieval settings of \cref{sec:exp-retrieval} differ and examine why \texttt{full} lands between the two single sources.
All comparisons except those by domain use the 233 questions that every model answered in all four settings, and every interval is a 95\% interval from a paired bootstrap over the compared questions.

\paragraph{Differences between settings.}
\Cref{tab:setting-diffs} gives the difference in BI between each pair of retrieval settings for every model.
\texttt{online-filtered} scores higher than \texttt{archive} for all eight models, by 0.92~BI on average.
If each sign were a fair coin, all eight would agree with probability 0.008.
For single models, however, the interval excludes zero only for GPT-6 Sol.
No interval of \texttt{full} against either single source excludes zero, for a single model or for the mean over models.

\begin{table}[!ht]
  \centering
  \footnotesize
  \caption{Difference in BI between retrieval settings on the 233 common questions, with 95\% intervals from a paired bootstrap with 4{,}000 resamples; positive values favor the first setting.}
  \label{tab:setting-diffs}
  \setlength{\tabcolsep}{5pt}
  \begin{tabular}{lccc}
    \toprule
    Model & \colhead{\texttt{online-filtered}\\$-$ \texttt{archive}} & \colhead{\texttt{full}\\$-$ \texttt{archive}} & \colhead{\texttt{full}\\$-$ \texttt{online-filtered}} \\
    \midrule
    GPT-6 Astra & +1.08 [$-$0.14, +2.31] & +0.57 [$-$0.46, +1.62] & $-$0.51 [$-$1.45, +0.33] \\
    DeepSeek V4.1 Flash & +0.90 [$-$0.90, +2.79] & +0.67 [$-$0.91, +2.22] & $-$0.23 [$-$1.66, +1.20] \\
    GPT-6 Sol & +1.50 [+0.15, +2.86] & +0.84 [$-$0.23, +1.92] & $-$0.66 [$-$1.99, +0.61] \\
    Claude Opus~5.5 & +0.88 [$-$0.40, +2.20] & +0.90 [$-$0.35, +2.18] & +0.02 [$-$1.29, +1.35] \\
    Kimi~K3 & +0.90 [$-$1.03, +2.80] & +0.69 [$-$0.92, +2.41] & $-$0.21 [$-$1.86, +1.40] \\
    GLM~5.3 & +0.81 [$-$0.72, +2.40] & +0.82 [$-$0.47, +2.21] & +0.01 [$-$1.44, +1.40] \\
    Qwen3.8-27B & +0.46 [$-$1.48, +2.50] & $-$0.56 [$-$2.50, +1.45] & $-$1.02 [$-$3.12, +1.04] \\
    Qwen3.6-35B-A3B & +0.88 [$-$1.78, +3.40] & +1.26 [$-$1.50, +3.95] & +0.38 [$-$2.23, +2.99] \\
    \midrule
    Mean over models & +0.92 [+0.12, +1.77] & +0.65 [$-$0.14, +1.46] & $-$0.28 [$-$0.91, +0.34] \\
    \bottomrule
  \end{tabular}
\end{table}

\begin{figure}[!ht]
  \centering
  \expfig{\linewidth}{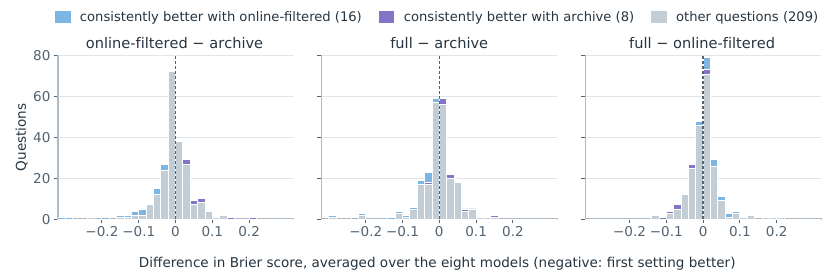}
  \caption{Difference in Brier score between two retrieval settings for each of the 233 common questions, averaged over the eight models; bins are 0.02 wide, and negative values favor the first setting.
  Colors mark, in all three panels, the questions consistently better under \texttt{online-filtered} or \texttt{archive} in the first panel; 110, 118 and 127 questions lie within 0.02 of zero in the three panels.}
  \label{fig:setting-diffs}
\end{figure}

\paragraph{Questions that favor one source.}
To see which questions drive these differences, \cref{fig:setting-diffs} shows, for each question, the difference in Brier score between two settings averaged over the eight models.
We call a question consistently better under one of two settings when a sign-flip test over the eight models gives $p<0.05$ and the mean difference in Brier score is at least 0.02.
By this rule, 16 questions are consistently better under \texttt{online-filtered} and 8 under \texttt{archive}.
To estimate what chance gives, we flip the sign of each model's difference at random, 4{,}000 times.
The rule then selects 3.8 questions for \texttt{online-filtered} and 3.9 for \texttt{archive} on average; 16 or more favor \texttt{online-filtered} in none of the simulations, and 8 or more favor \texttt{archive} in 4.9\% of them.
Web search thus helps on more questions than chance would give, the archive on a few at the edge of chance, and the other 209 questions show no consistent difference.
No single question remains significant after a Benjamini--Hochberg correction at a false discovery rate of 0.05, 0.10 or 0.20.
The two lists therefore give examples of each kind, without establishing any single question.
Of the 16 questions that favor web search, six concern finance, five technology, two culture, two geopolitics and one the economy; the 8 that favor the archive concern technology, the economy, finance and sports.
The first group includes whether Ariane~6 launches at least twice, whether Kimi~K3 is released by the end of July, Intel's quarterly revenue and the Emmy for Outstanding Drama Series.
The second includes an ISRO test flight, the rate decisions of the central banks of Norway and Australia, and the vehicle deliveries of Tesla and Li Auto.

\paragraph{How \texttt{full} combines the sources.}
For each model, we regress its \texttt{full} probability on its \texttt{archive} and \texttt{online-filtered} probabilities across the 233 questions.
The weight on \texttt{archive} averages 0.44 and ranges from 0.32 to 0.57 across models, and the weight on \texttt{online-filtered} averages 0.50 and ranges from 0.38 to 0.67.
Nearly equal weights are expected both if agents blend the two sources and if \texttt{full} were a third run with the same information.
To tell these apart, we take the 1{,}308 model--question pairs in which the three Brier scores all differ.
\texttt{full} is better than both single sources in 32\% of them, between them in 36\% and worse than both in 32\%, close to the one third each that a third independent run would give.
Agents given both sources thus behave much like a third run, and they do not pick out the better source.
Averaging a model's \texttt{archive} and \texttt{online-filtered} probabilities directly scores at least as well as \texttt{full} for six of the eight models (\cref{tab:setting-average}), although this average draws on two runs per question.

\begin{table}[!ht]
  \centering
  \footnotesize
  \caption{BI on the 233 common questions of each retrieval setting and of the average of each model's \texttt{archive} and \texttt{online-filtered} probabilities.}
  \label{tab:setting-average}
  \setlength{\tabcolsep}{5pt}
  \begin{tabular}{lcccc}
    \toprule
    Model & \texttt{archive} & \texttt{online-filtered} & \texttt{full} & Average of the two \\
    \midrule
    GPT-6 Astra & 57.84 & 58.91 & 58.40 & 58.62 \\
    DeepSeek V4.1 Flash & 57.80 & 58.69 & 58.46 & 58.73 \\
    GPT-6 Sol & 56.61 & 58.10 & 57.45 & 57.64 \\
    Claude Opus~5.5 & 56.17 & 57.05 & 57.07 & 56.85 \\
    Kimi~K3 & 56.78 & 57.68 & 57.47 & 57.74 \\
    GLM~5.3 & 55.93 & 56.74 & 56.76 & 56.68 \\
    Qwen3.8-27B & 55.24 & 55.70 & 54.68 & 56.06 \\
    Qwen3.6-35B-A3B & 50.84 & 51.72 & 52.09 & 52.33 \\
    \bottomrule
  \end{tabular}
\end{table}

\paragraph{Domains.}
\Cref{tab:setting-domains} gives the gains of \cref{fig:domains} averaged over models, with \texttt{online-filtered} added, and the differences between settings with intervals over the questions of each domain.
Every difference between \texttt{full} and \texttt{archive} has an interval that includes zero.
Between \texttt{online-filtered} and \texttt{archive}, only Health has an interval that excludes zero.

\begin{table}[!ht]
  \centering
  \footnotesize
  \caption{Gain in BI over \texttt{none} by domain, averaged over models, and differences between settings with 95\% intervals over the questions of the domain.
  Each model uses the questions of the domain that it answered in both compared settings, as in \cref{fig:domains}, and the intervals use 1{,}500 resamples; columns headed $>0$ count the models with a positive difference.}
  \label{tab:setting-domains}
  \setlength{\tabcolsep}{2.6pt}
  \begin{tabular}{lrccccccc}
    \toprule
    Domain & $n$ & \texttt{archive} & \colhead{\texttt{online-}\\\texttt{filtered}} & \texttt{full} & \colhead{\texttt{full}\\$-$ \texttt{archive}} & $>0$ & \colhead{\texttt{online-filtered}\\$-$ \texttt{archive}} & $>0$ \\
    \midrule
    Finance & 77 & +4.47 & +3.87 & +4.31 & $-$0.12 [$-$1.25, +0.96] & 3 & $-$0.43 [$-$1.55, +0.88] & 3 \\
    Economy & 65 & +3.69 & +3.39 & +2.67 & $-$0.86 [$-$1.85, +0.20] & 3 & $-$0.08 [$-$1.32, +1.19] & 4 \\
    Technology & 27 & +3.53 & +6.91 & +6.31 & +2.35 [$-$0.58, +5.71] & 7 & +2.96 [$-$0.74, +7.42] & 6 \\
    Health & 24 & +2.30 & +6.24 & +4.04 & +1.74 [$-$1.17, +4.69] & 6 & +3.92 [+1.43, +6.29] & 7 \\
    Sports & 17 & +1.07 & $-$1.02 & +0.55 & $-$0.53 [$-$2.90, +1.71] & 3 & $-$1.79 [$-$4.22, +0.33] & 2 \\
    Culture & 16 & +1.66 & +5.98 & +5.39 & +3.51 [$-$0.08, +7.80] & 8 & +4.22 [$-$0.03, +9.12] & 8 \\
    Geopolitics & 15 & $-$2.26 & $-$0.70 & $-$0.70 & +1.71 [$-$2.13, +6.33] & 6 & +1.71 [$-$1.64, +5.50] & 5 \\
    Weather & 13 & +1.29 & +3.47 & +4.32 & +2.55 [$-$0.27, +5.16] & 6 & +2.14 [$-$0.10, +4.57] & 5 \\
    Politics & 33 & +0.40 & +1.53 & +2.09 & +1.31 [$-$0.90, +4.23] & 5 & +1.76 [$-$0.45, +4.12] & 7 \\
    \bottomrule
  \end{tabular}
\end{table}

\clearpage
\section{Failure Analysis Details}
\label{app:failures}

To find failures on individual questions, we classify each question by whether its Brier score, averaged over models, exceeds 0.25, the score of the constant forecast, without retrieval and with \texttt{full}.
Retrieval corrects 41 questions and misleads 11.
Another 76, about a quarter of the benchmark, stay above 0.25 in both settings; on 28 of them, no model beats the constant forecast.
We traced eight such questions in an earlier run of the same settings and describe one case of each kind; the probabilities below come from the final runs.

\paragraph{Retrieval corrects outdated priors.}
On whether the Bank of Korea would hold its base rate at 2.50\% on July~16, models without tools extrapolated a long series of holds and assigned 0.60 to 0.83 to a hold.
Retrieval surfaced the committee's split vote, the governor's guidance and accelerating inflation.
Every model then moved toward the hike that followed, to at most 0.30 in every retrieval setting.

\paragraph{Unofficial reports can override newer official statements.}
On whether ISRO would launch the uncrewed Gaganyaan G1 mission by September~15, every model assigns at most 0.40 to a launch under \texttt{archive}, and no launch occurred.
Under \texttt{online-filtered}, four models rise to between 0.50 and 0.75.
DS V4.1 Flash and Kimi~K3 cite a ground-station window relayed on social media, which the news archive lacked, and let it outweigh the program head's more recent guidance of a launch at the end of 2026 or later.
Under \texttt{full}, which also searches the archive, only the two Qwen models remain above 0.30.
This case shows one way retrieval misleads, not the most common one: 7 of the 11 misled questions concern finance.

\paragraph{A public consensus can itself be wrong.}
On whether Disney's live-action \emph{Moana} would gross more than \$56.6 million on its opening weekend, models assign 0.72 to 0.99 in every setting.
Retrieval returned an industry consensus well above the threshold, and in the audited traces the models that also found bearish forecasts treated them as tail risk.
The film opened at \$43.1 million, below every published forecast.

\paragraph{Correct evidence can be applied to the wrong case.}
On whether the U.S. Mint would release a coin bearing Donald Trump's likeness by September~15, retrieval lowers every model's probability in every setting, to at most 0.35 under \texttt{full}.
In the audited traces, the models applied the multi-month production timeline of a gold coin to every Trump coin.
They missed a \$1 coin that was released on an unannounced fast track.

\section{Limitations}
\label{app:limitations}

Our evaluation has two main limitations.

First, the two information sources trade coverage against guarantees.
The archive cannot leak and returns the same results in every run.
It holds only English news captured by CC-News and the Wikipedia revisions on the questions' entities and topics, however, so it lacks many company, regulator and agency pages.
Screened web search reaches such pages, but it relies on an LLM reviewer (\cref{sec:env-web}), and it cannot be repeated exactly, because a live search engine may stop returning some pages and its results change over time.
The reviewer can also miss later information, although such misses were rare when we reviewed pages again at a higher reasoning effort (\cref{app:environment}).
Because the archive cannot leak, such misses appear to account for at most the small lead of \texttt{online-filtered} over \texttt{archive}, a lead that also includes what only the web reaches (\cref{sec:exp-retrieval}).
Extending the archive to more kinds of sources would bring more of what the web reaches under the guarantees of the archive.

Second, the first release supports aggregate comparisons better than fine-grained ones.
All 287 questions are posed at one forecast date, and each model forecasts each question once per setting.
For a single model, the differences between settings therefore remain within sampling uncertainty, and so do most differences between settings within a domain (\cref{app:settings}).
Question sets at further forecast dates, which the workflow already produces (\cref{app:dates}), and repeated runs would reduce this uncertainty.

\end{document}